%% file: main.tex
\documentclass{article}

\PassOptionsToPackage{numbers}{natbib}
\usepackage[preprint]{neurips_2026}

\usepackage[utf8]{inputenc} 
\usepackage[T1]{fontenc}    
\usepackage{hyperref}       
\usepackage{url}            
\usepackage{booktabs}       
\usepackage{amsfonts}       
\usepackage{nicefrac}       
\usepackage{microtype}      
\usepackage{xcolor}         

\usepackage{graphicx}
\usepackage{subcaption}

\usepackage{xspace}
\usepackage{amsmath}
\usepackage{multirow}
\usepackage{algorithm}
\usepackage{algorithmic}
\usepackage{wrapfig}
\usepackage{fancybox}
\usepackage[most]{tcolorbox}

\usepackage{comment}
\excludecomment{blockcomment}

\usepackage{amssymb}
\usepackage{mathtools}
\usepackage{amsthm}
\usepackage[capitalize,noabbrev]{cleveref}

\title{SpinDoctor: Finding Innocuous Prompts 
  for Biased Image Generation}

\author{%
  Anna Gerchanovsky \\
  Duke University\\
  \texttt{anna.gerchanovsky@duke.edu} \\
  \And
  Omer Akgul \\
  RSAC Labs \\
  \texttt{omer.akgul@onersac.com} \\
  \And
  Lujo Bauer \\
  Carnegie Mellon University\\
  \texttt{lbauer@cmu.edu} \\
  \And
  Michael K. Reiter \\
  Duke University\\
  \texttt{michael.reiter@duke.edu} \\
}

\begin{document}

\maketitle

\input{macros}

\begin{abstract}
    We introduce \SysName, a method for discovering innocuous prompts 
    for text-to-image models that result in images with a chosen target bias, 
    while retaining the quality of the images and prompts. 
    These prompts present a risk for user manipulation, 
    in that a user who is given these seemingly unbiased prompts might be willing 
    to use them and would receive biased images in return. 
    Our method is flexible: it uses models in a black-box way only and the target 
    image bias can easily be changed, and so our method can be applied 
    with various motivations and in various scenarios. 
    Our method shows some transferability between models---we 
    saw the same bias emerge in 78.6\% of cases---such 
    that a victim who uses a prompt generated via our method using one model still 
    likely receives a biased image from another model. In our experiments, 
    a subset of which we corroborated with human evaluation, 
    we were able to find an innocuous prompt that moved generated images in 
    the direction of the desired bias in all but two of 304 experiments, 
    with an average improvement of 34.8\% and 33.5\% on two open-source
    text-to-image models.
\end{abstract}

\input{intro}
\input{related}
\input{threatmodel}
\input{methods}
\input{evalresults}
\input{discussion}
\input{conclusion}

\bibliography{full, refs}
\bibliographystyle{abbrvnat}

\input{appendix}


\end{document}

%% file: macros.tex
\definecolor{hazel}{RGB}{142,118,24} 
\definecolor{aigreen}{RGB}{0,153,24} 
\definecolor{brown}{RGB}{165,42,42}

\newif\ifdraft
\drafttrue
\draftfalse

\ifdraft 
    \newcommand{\todo}[1]{\textcolor{blue}{TODO: #1}}
    \newcommand{\note}[1]{\textcolor{red}{NOTE: #1}}
    \newcommand{\mkr}[1]{\textcolor{purple}{MKR: #1}}
    \newcommand{\omer}[1]{\textcolor{hazel}{Omer: #1}}
    \newcommand{\lujo}[1]{\textcolor{brown}{Lujo: #1}}
    \newcommand{\anna}[1]{\textcolor{magenta}{Anna: #1}}
    \newcommand{\new}[1]{\textcolor{magenta}{New: #1}}
\else 
    \newcommand{\todo}[1]{}
    \newcommand{\note}[1]{}
    \newcommand{\mkr}[1]{}
    \newcommand{\omer}[1]{}
    \newcommand{\lujo}[1]{}
    \newcommand{\anna}[1]{}
    \newcommand{\new}[1]{#1}
\fi

\newcommand{\SysName}{SpinDoctor\xspace}
\newcommand{\sysName}{SpinDoctor\xspace}
\newcommand{\pdouble}{\ensuremath{\makebox[4pt][l]{\shortstack{\scriptsize $\ast$\\[-2pt]\scriptsize $\ast$}}}}

\newtcolorbox[auto counter]{takeawaybox}{
    boxrule = 1pt,
    colframe = black,
    rounded corners,
    arc = 5pt,
    boxsep = 0pt,
    left = 4pt,
    right = 4pt,
    top = 4pt,
    bottom = 4pt,
    colback = white,
}

\newcommand{\takeaways}[1]{
    \begin{takeawaybox}
        \textbf{Takeaway ~\thetcbcounter:} #1
    \end{takeawaybox}
}

\NewDocumentCommand{\pval}{}{\ensuremath{p}\xspace}
\NewDocumentCommand{\basePrompt}{}{\ensuremath{p_{\mathsf{b}}}\xspace}
\NewDocumentCommand{\manipulatedPrompt}{}{\ensuremath{p_{\mathsf{m}}}\xspace}
\NewDocumentCommand{\imagePolicy}{}{\ensuremath{Q_{\mathsf{im}}}\xspace}
\NewDocumentCommand{\imagePolicyWeights}{}{\ensuremath{W_{\mathsf{im}}}\xspace}
\NewDocumentCommand{\promptPolicyPos}{}{\ensuremath{Q_{\mathsf{+}}}\xspace}
\NewDocumentCommand{\promptPolicyNeg}{}{\ensuremath{Q_{\mathsf{-}}}\xspace}
\NewDocumentCommand{\imageQuestion}{}{\ensuremath{q_{\mathsf{im}}}\xspace}
\NewDocumentCommand{\imageQuestionWeight}{}{\ensuremath{w_{\mathsf{im}}}\xspace}
\NewDocumentCommand{\promptQuestion}{}{\ensuremath{q_{\mathsf{pr}}}\xspace}
\NewDocumentCommand{\cutoffMin}{}{\ensuremath{c_{\mathsf{min}}}\xspace}
\NewDocumentCommand{\cutoffMax}{}{\ensuremath{c_{\mathsf{max}}}\xspace}
\NewDocumentCommand{\scorePromptFunc}{o o}{\ensuremath{\mathit{judge_{\mathsf{pr}}}\IfNoValueF{#1}{\mathopen{(}{#1}\IfNoValueF{#2}{,\,{#2}}\mathclose{)}}}\xspace}
\NewDocumentCommand{\scoreImageFunc}{o o}{\ensuremath{\mathit{judge_{\mathsf{im}}}\IfNoValueF{#1}{\mathopen{(}{#1}\IfNoValueF{#2}{,\,{#2}}\mathclose{)}}}\xspace}
\NewDocumentCommand{\scorePromptFuncName}{}{\ensuremath{\mathit{judge_{\mathsf{pr}}}}\xspace}
\NewDocumentCommand{\scoreImageFuncName}{}{\ensuremath{\mathit{judge_{\mathsf{im}}}}\xspace}
\NewDocumentCommand{\generateImage}{o o}{\ensuremath{\mathit{gen_{\mathsf{im}}}\IfNoValueF{#1}{\mathopen{(}{#1}\IfNoValueF{#2}{,\,{#2}}\mathclose{)}}}\xspace}
\NewDocumentCommand{\minFunc}{o o}{\ensuremath{\mathit{min}\IfNoValueF{#1}{\left({#1}\IfNoValueF{#2}{,\,{#2}}\right)}}\xspace}

\NewDocumentCommand{\penaltyFactor}{}{\ensuremath{f}\xspace}
\NewDocumentCommand{\penaltyFunc}{o}{\ensuremath{\mathit{F}\IfNoValueF{#1}{\left({#1}\right)}}\xspace}
\NewDocumentCommand{\imScore}{}{\ensuremath{s_{\mathsf{im}}}\xspace}
\NewDocumentCommand{\score}{}{\ensuremath{s_{\mathsf{tot}}}\xspace}
\NewDocumentCommand{\disqualified}{}{\ensuremath{d}\xspace}
\NewDocumentCommand{\promptQuestionType}{}{\ensuremath{t_{\mathsf{q}}}\xspace}
\NewDocumentCommand{\isGood}{}{\ensuremath{pos}\xspace}
\NewDocumentCommand{\imagesBase}{}{\ensuremath{I_{\mathsf{b}}}\xspace}
\NewDocumentCommand{\imageBase}{}{\ensuremath{i_{\mathsf{b}}}\xspace}
\NewDocumentCommand{\imagesManipulated}{}{\ensuremath{I_{\mathsf{m}}}\xspace}
\NewDocumentCommand{\imageManipulated}{}{\ensuremath{i_{\mathsf{m}}}\xspace}
\NewDocumentCommand{\seed}{}{\ensuremath{\mathit{seed}}\xspace}
\NewDocumentCommand{\numSeeds}{}{\ensuremath{N}\xspace}

\NewDocumentCommand{\postoks}{}{\ensuremath{tokens_{\mathsf{affirmative}}}\xspace}
\NewDocumentCommand{\negtoks}{}{\ensuremath{tokens_{\mathsf{negative}}}\xspace}
\NewDocumentCommand{\onehot}{o}{\ensuremath{\mathit{onehot}\IfNoValueF{#1}{\left({#1}\right)}}\xspace}
\NewDocumentCommand{\modelcall}{o o}{\ensuremath{\mathit{VL\_model}\IfNoValueF{#1}{\mathopen{(}{#1}\IfNoValueF{#2}{,\,{#2}}\mathclose{)}}}\xspace}
\NewDocumentCommand{\modelname}{}{\ensuremath{\mathit{VL\_model}}\xspace}
\NewDocumentCommand{\probs}{}{\ensuremath{probs}\xspace}
\NewDocumentCommand{\posprobs}{}{\ensuremath{probs_{\mathsf{affirmative}}}\xspace}
\NewDocumentCommand{\softmax}{o}{\ensuremath{\mathit{softmax}\IfNoValueF{#1}{\left({#1}\right)}}\xspace}

\NewDocumentCommand{\iter}{}{\ensuremath{i}\xspace}
\NewDocumentCommand{\maxiter}{}{\ensuremath{N}\xspace}
\NewDocumentCommand{\allprompts}{}{\ensuremath{P_{\mathsf{all}}}\xspace}
\NewDocumentCommand{\prompt}{}{\ensuremath{p}\xspace}
\NewDocumentCommand{\bestprompts}{}{\ensuremath{P_{\mathsf{best}}}\xspace}
\NewDocumentCommand{\bestprompt}{}{\ensuremath{p_{\mathsf{best}}}\xspace}
\NewDocumentCommand{\iterprompts}{}{\ensuremath{P_{\mathsf{iter}}}\xspace}
\NewDocumentCommand{\paraphrase}{o}{\ensuremath{\mathit{paraphrase}\IfNoValueF{#1}{\left({#1}\right)}}\xspace}
\NewDocumentCommand{\overallJudgeFuncName}{}{\ensuremath{\mathit{score}}\xspace}
\NewDocumentCommand{\overallJudgeFunc}{o o}{\ensuremath{\mathit{score}\IfNoValueF{#1}{\mathopen{(}{#1}\IfNoValueF{#2}{,\,{#2}}\mathclose{)}}}\xspace}
\NewDocumentCommand{\numbeams}{}{\ensuremath{k}\xspace}
\NewDocumentCommand{\scoresort}{o}{\ensuremath{\mathit{sorted}\IfNoValueF{#1}{\left({#1}\right)}}\xspace}
\NewDocumentCommand{\iterScore}{}{\ensuremath{s_{\mathsf{iter}}}\xspace}
\NewDocumentCommand{\bestScore}{}{\ensuremath{s_{\mathsf{best}}}\xspace}
\NewDocumentCommand{\bestUnused}{o}{\ensuremath{\mathit{get\_best\_unused}\IfNoValueF{#1}{\left({#1}\right)}}\xspace}

\NewDocumentCommand{\imageScoresManipulated}{}{\ensuremath{\mathit{MIS}}\xspace}
\NewDocumentCommand{\imageScoresBase}{}{\ensuremath{\mathit{BIS}}\xspace}
\NewDocumentCommand{\imageScoresDelta}{}{\ensuremath{\Delta \mathit{IS}}\xspace}
\NewDocumentCommand{\imageScoresRatio}{}{\ensuremath{\frac{\imageScoresManipulated}{\imageScoresBase}}\xspace}

%% file: intro.tex
\section{Introduction}
\label{sec:intro}

Writing prompts for generative AI models can on the surface seem simple,
yet in practice users 
must translate a vague intent into concrete 
constraints, style cues, and model-specific conventions, 
and then iterate when the model behaves unexpectedly. 
Prior work shows that non-experts often struggle to 
improve prompts systematically and instead rely on 
ad-hoc trial-and-error~\cite{zamfirescu2023johnny}. 
This friction can be even worse for text-to-image 
generation: mapping words to images is inherently 
underdetermined, users might need to name fine-grained 
visual attributes (e.g., composition),
and it can be unclear whether a prompt revision 
truly improved the outcome, leading to long, iterative 
prompt journeys~\cite{goloujeh2024promptjourney}.

To reduce this burden, many systems 
provide \emph{prompt assistance}---automatic prompt rewriting, 
suggestion, and optimization---to elicit better outputs 
with less user effort, especially for text-to-image tools. 
This design choice is both widely deployed 
(e.g., DALL\textperiodcentered E~3 via ChatGPT, Midjourney's \texttt{/describe}, and Copilot's Imagine)~\cite{openai2023dalle3,midjourney2025describe,microsoft_copilot_imagine} and actively studied in the research literature~\cite{feng2023promptmagician,chen2024tailoredvisions}.

However, recent work shows that prompt-assistance 
paradigms can be manipulated with nearly imperceptible 
bias placements.
\citet{lin2025llm} show that users may accept seemingly 
helpful rewrites (e.g., synonym replacements) 
while the downstream model behavior is 
steered in an attacker-chosen 
direction (e.g., brand name mention).
While this issue has been studied primarily for 
LLMs, the same threat may apply
to other modalities, such as image 
generation.
In image generation, objective functions for success
that do not require user input are
harder to define. \citet{lin2025llm} use the presence of specific 
strings to judge success, as does \citet{Arawjo_2024_chainforge}. 
Prompt optimizers use benchmark datasets
to evaluate success~\cite{cui2024phaseevo,
fernando2024PromptBreeder,yang2024large}.
In order to evaluate optimized image-generation prompts,
\citet{Li2024PromptTist} fine-tunes a model specifically for the task of 
assessing image quality---which is both more difficult and less flexible
than methods used for text generation.

If feasible, such attacks are especially 
concerning for text-to-image systems. 
First, biases in model-produced images may
be difficult to 
detect: bias can surface in what is depicted, the roles of individuals in images,
their expressions, clothing, and other contextual signals~\cite{naik2023socialbiases,chinchure2024tibet,raza2025promptingaway,huang2025implicitbiasinjection}.
Second, biased visual framings may be especially 
impactful because images are often more 
memorable than text~\cite{standing1973learning10000},
and can exert outsized effects on credibility 
judgments relative to purely verbal 
descriptions~\cite{newman2024}.
Pairing images with accompanying text can further 
amplify their persuasive impact~\cite{newman2012truthiness}.

Finally, because prompting for text-to-image is often 
more difficult
than prompting 
LLMs~\cite{zamfirescu2023johnny,goloujeh2024promptjourney}, 
users and platforms increasingly rely on 
assistance (e.g.,~\citet{microsoft_copilot_imagine}),
increasing the impact of attacks that target the assistance layer.
This highlights prompt suggestion in text-to-image models as an important 
attack surface that warrants a deeper investigation.

\begin{figure}
    \centering
    \includegraphics[width=0.65\linewidth]{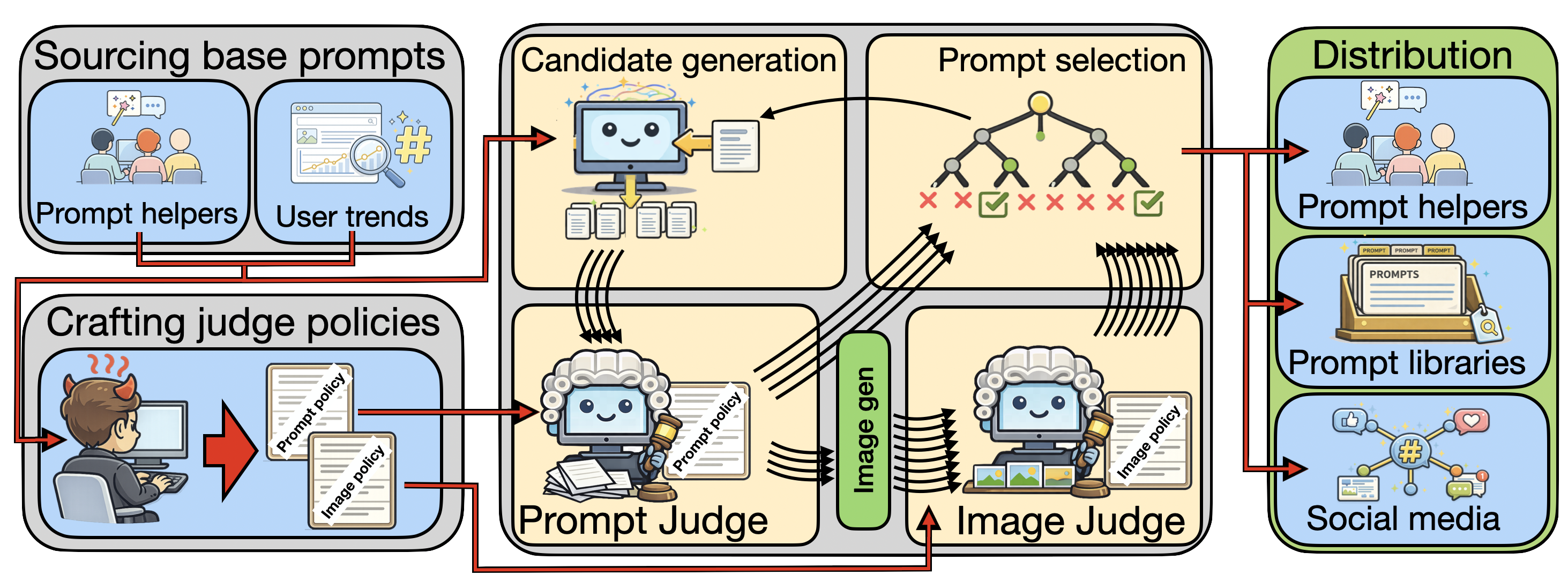}
    \caption{Overview of \SysName. Sourced prompts and the manipulator 
    policy are fed into \SysName for iterative optimization. 
    The resulting prompts are distributed to victims through prompt helpers, 
    libraries, and social forums. }
    \label{fig:overview}
\end{figure}

In this work, we investigate whether it is possible to
modify image-generation model prompts so that images generated with
those prompts exhibit properties of a
manipulator's choosing (e.g., a country's flag, 
image subject gender, appearance of a soda brand like Sprite); 
at the same time, the modified prompts themselves
must appear innocuous to users and not exhibit the properties induced
in the images.
As shown in \citet{guo2025PrompTHis}, the difference in prompts 
can differ from the difference in the generated images---indicating
that a prompt altered without adding a particular bias can still 
include this bias in the generated image.
We show such manipulations are possible for 
text-to-image generation through an approach
we call \emph{Stealthy Prompt-Induced Nudging} (Spin).
We require only black-box access and a 
natural-language policy (with optional weights) 
specifying desired prompt and image properties; 
\SysName then optimizes for innocuous prompts 
that induce those properties.

Inspired by prior work~\cite{naseh2025backdooring, huang2025implicitbiasinjection, bianchi2023demographicbias}, 
we evaluate on a seven category dataset---which
define the subject matter of the image---with 
  multiple target biases
  per category (e.g.,
gender of the image subject and
the brand featured).
We test on Stable Diffusion and evaluate
prompts and images with Qwen.
On average, \SysName achieves
$>30\%$ improvement toward the target bias,
with the best examples
(``politician’’ $\rightarrow$ female politician) improving
by 64\% on average
(i.e., ``female politician'' 
appearing 64\% more frequently compared to base,
see~\cref{fig:appendix:attack:success_example:politician_female}).
Images from modified prompts exhibit the target 
properties with minimal prompt- and image-quality change.
We corroborate results for manipulation success, image 
and prompt quality through a blind human evaluation.
Finally, we evaluate transferability by testing 
manipulated prompts crafted for Stable Diffusion
on closed-source models, where we achieve success in 11 of 14 cases.

Paper organization is as follows: 
related work (\cref{sec:related}); 
attack and threat model (\cref{sec:threat}); 
\SysName details (\cref{sec:methods}); 
evaluation and results (\cref{sec:evalres});
discussion and limitations (\cref{sec:discussion});
and conclusion in \cref{sec:conclusion}.

%% file: related.tex
\section{Related work}
\label{sec:related}

\subsection{Bias in LLMs and text-to-image generation}

Studies of bias in generative systems build on a longer line of
research on bias in computer systems.  Design choices
embed social biases in systems, leading to measurable bias in those
systems~\cite{friedman_nissenbaum_1996_bias}.

In the LLM setting, \citet{bender_etal_2021_stochastic} argue that
web-trained language models reproduce harms present in the underlying
data.  \citet{gallegos-etal-2024-bias} consolidate recent evaluation
and mitigation approaches for bias and fairness in LLMs, providing
taxonomies of metrics, datasets, and interventions across the model
lifecycle.
Recent work has also focused on text-to-image models. 
\citet{Garcia_2023_CVPR} annotate and analyze 
image datasets, showing that bias persists in 
text-to-image generation. 
\citet{luccioni_2023_neurips} propose an evaluation 
framework for diffusion 
models that measures representational 
(demographics, etc.) variation.
\citet{seshadri-etal-2024-bias} examine 
expression of bias 
in Stable Diffusion and show that naive
bias measurements might be flawed, 
in that models might show bias 
when prompted with generic benchmarking 
prompts but act less biased with 
caption-like prompts.
On the mitigation 
side, \citet{Zhang_2023_ICCV} introduce ITI-GEN, 
which uses reference images and learned prompt 
embeddings to encourage more uniform attribute 
coverage without fine-tuning. 
Later work also explores inference-time or 
training-free debiasing 
strategies \cite{Kim_2025_CVPR,Park_2025_ICCV}.

\subsection{Optimizing prompts for image generation}

Optimizing prompts for increased model 
performance isn't new. 
This has previously been explored in the text 
domain~\cite{shin2020autoprompt,mishra2025promptaid,
yang2024large,fernando2024PromptBreeder,cui2024phaseevo},
including optimization for preference~\cite{Wang2026Promptimizer,
gao_2024_align}.
Research has since expanded to image 
generation. 
Several works use reinforcement 
learning to create 
a prompt rewriter, crafting higher performing 
prompts \cite{hao2023optimizing, chen2024tailoredvisions,
guo2025PrompTHis,Wang2024PromptCharm,brade2023promptify}. 
Related to our work, \citet{xiang2025promptsculptor} 
and \citet{chen2025t2i} define 
agents to judge produced images and 
suggest feedback for prompt 
rewriting. Since generated images are 
benign, desired features can be 
declared in the prompt.

Our work differs from others in important ways. 
(i) Our goal is a double optimization: prompts must remain 
innocuous but resulting images biased. This means our objective 
cannot be simply stated in the prompt. 
(ii) We use off-the-shelf LVLM 
judges for tasks 
that the LVLM judges could 
refuse (bias injection) if the objective was 
explicit as with prior work~\cite{xiang2025promptsculptor}. 
This design also allows arbitrary
swaps of judges for more advanced ones.
(iii) Our optimization is flexible: the manipulator 
can specify biases and adjust the importance of 
them through weights.

%% file: threatmodel.tex
\section{Threat model and attack overview}
\label{sec:threat}

\paragraph{Victims}
Our threat model assumes that users are willing to use 
prompts they haven't authored, perhaps due to 
prompt writing friction~\cite{goloujeh2024promptjourney}.
Sources for these prompts include:
prompt optimizers and 
helpers~\cite{brade2023promptify,mishra2025promptaid,
su2023promptmindrefinepersonalized,Wang2024PromptCharm,feng2023promptmagician} 
that suggest modifications to prompts, 
AI assistants suggesting prompts 
in-conversation~\cite{lin2025llm}, prompt libraries 
or online articles~\cite{wind2023prompts} that curate prompts, etc.

\paragraph{Manipulator} A manipulator with access 
to a pipeline that serves prompts to 
users aims to manipulate suggestions such that
(i) the user doesn't suspect bias, 
(ii) manipulated and original prompt quality is comparable,
(iii) when the prompt is used in image generation, 
images contain a bias of the manipulator's choosing. 
\SysName expects manipulators to specify a set of traits for the modified
prompt, which we call the \emph{prompt policy}; as well as traits for generated images in 
response to the prompts, which we call the \emph{image policy}.

\paragraph{Base prompts}
The manipulation scheme we present relies on iterating on prompts 
until manipulator objectives (prompt and image policy) are met. 
We call the first version of the prompt the \emph{base prompt}.
The manipulator may source the base prompt directly from victims (e.g.,
when the victim is submitting a prompt to a prompt optimizer); or they
may generate the base prompt themselves (e.g., when the user is only
going to see the manipulated prompt). In the latter case, starting from
an innocuous base prompt may make it easier for an attacker to produce
a manipulated prompt that does not, on inspection, appear to have been
manipulated.

\paragraph{Attack flow} Once base prompts are sourced 
and the manipulator policies specified, the prompts and policies are 
passed to \SysName. \SysName optimizes the base prompts on an image generation model
according to policy and outputs innocuous manipulated prompts
that, when used on text-to-image models, 
lead to images consistent with the manipulator's
image policy.
Finally, the manipulated prompt is distributed to victims, e.g., 
prompt optimizers might 
suggest the manipulated prompt, or prompts might be posted on prompt libraries. 
\cref{fig:overview} illustrates this flow.

%% file: methods.tex
\section{\SysName}
\label{sec:methods}

\SysName 
consists of three main components, each responsible for one of these tasks: (i) generating candidate prompts, (ii) evaluating candidate prompts and resulting images 
with policy-adhering judges, and (iii) selecting
successful candidates and iterating.
In each iteration, the best prompts from the previous iteration are paraphrased, resulting in a
new batch of prompts. These prompts are judged, potentially incurring
a penalty factor, and
then used to generate images. The images are in turn judged on how well
they convey the target bias, and these scores are combined with the penalty
factor to select new set of best prompts, which are then used by the next iteration.

We evaluated \SysName on Stable Diffusion 3.5 
(medium)~\cite{stabilityai2026stablediffusion}
and FLUX.2 [klein] 4B~\cite{blackforestlabs2026flux}.
A subset of experiments was also performed on 
GPT Image 1.5~\cite{openai2025gptimage1},
as described in \cref{sec:evalres:success:results}.

\subsection{Individual policy judgements}
\label{sec:method:individual}
While the prompt and image judges 
could have various implementations---image classifiers, 
embeddings, color analysis, 
etc.---we use a large vision language model,
instruction-tuned Qwen 2.5 VL 7B~\cite{qwen2025blog}, 
to judge prompts and generated images 
on adherence to their respective policies.
We later validate this model's judgement by comparing
prompt scores and manipulation success 
with an alternative judge model, GPT 4.1, 
in \cref{sec:evalres:success:alt},
and with human judgement in \cref{sec:evalres:success:human}.

Each policy consists of a set of questions with yes or no answers,
and judgements are made
in the form of the probability that this question about the prompt or image
returns an affirmative answer from the VL model.
This is calculated as the collective probability of the next 
token being affirmative  (``Yes'', ``yes'', etc.)
over the token being negative (``No'', etc.),
as shown in more detail in \cref{alg:methods:score}.

\subsection{Overall prompt scores}
\label{sec:method:overall}
Our goal is for prompts to be high quality and innocuous 
and for the corresponding images to be high quality but biased. We achieve this via a 
two-tiered scoring process, combining judgments about the prompt 
and judgments about the images that result from that prompt.
The process of computing the
overall score of a prompt 
is shown in \cref{alg:methods:judge} in the appendix.
Image scores are calculated by averaging image judgements for all questions 
in the image policy over eight generated images.
The image policy 
can optionally include weights for the importance of each element in scoring.

Judgements on the prompt policy
 are used to discard prompts or apply a penalty
to the image score.
We divide the prompt policy
into positive (\promptPolicyPos) and negative attributes (\promptPolicyNeg).
\promptPolicyPos rewards prompts that maintain high quality and \promptPolicyNeg 
penalizes prompts for showing bias.
For a question about a positive attribute from the prompt policy,
we consider the ratio of the \textit{yes} 
probability for the \textit{base prompt}
to that of the \textit{manipulated prompt}, and vice versa 
for negative attributes.
If this ratio is above a cutoff, the prompt is 
disqualified\footnote{We use a cutoff of 4 for positive attributes and 
2 for negative attributes,
as we empirically determined these to perform well.
Without this cutoff,
these ratios would result in a penalty factor of 0.66 and 0.87,
respectively, or a 34 and 13\% reduction in the overall prompt score
when the image score is multiplied by the factor.
}. 
If the manipulated prompt is judged to be worse than the base prompt 
but below the cutoff,
the penalty factor \penaltyFactor is adjusted, by
multiplying \penaltyFactor by the result
of applying an increasing and concave function \penaltyFunc to this 
ratio\footnote{The 
function should increase so lower ratios result in a higher penalty,
and we choose our function to be concave to have diminishing
returns on ratios once they near 
one. We use $F(x) = \sqrt{1 - (x - 1)^2}$.
}.
The overall score of the prompt---which 
reflects the (extent of) presence of a desired image bias
and the quality of the prompt---is
the average image score multiplied by the
penalty factor, where a higher penalty factor is better.

\subsection{Iterative loop}
\label{sec:method:iterative}
The final prompt is produced iteratively, 
as shown in \cref{alg:attack:iter}.
\SysName begins by paraphrasing the base prompt.
Prompts are paraphrased with a BART paraphrasing model~\cite{Siow:2025:Bart}.
We use beam generation with the number of outputs equal to the 
number of beam groups to get more varied options. 
By default, we generate
10 paraphrases for each prompt but increase this by 10, up to a total of 30,
if needed.
Those candidate prompts are then judged (or discarded) 
by the \emph{prompt judge} and used
to generate images, which are judged by the \emph{image judge}.
Then, the highest scoring $k$ prompts, 
as described in \cref{sec:method:overall} are kept.
This loop repeats 10 times.
On every iteration, we paraphrase 
the $k$ best prompts from the previous iteration, 
evaluate all paraphrases,
and select the new $k$ best prompts to iterate on further. 
If too many prompts are discarded in any iteration, leaving fewer
than $k$ valid prompts,
more paraphrases are generated. 
If this does not produce $k$ or more valid prompts, \SysName draws new prompts from the 
pool of best past prompts 
that have not yet been paraphrased.
\SysName also draws from this pool if the best scores for an iteration
decrease in comparison to the previous iteration.
\SysName returns the single best prompt found across all iterations.

%% file: evalresults.tex
\section{Evaluation and results}
\label{sec:evalres}
In this section, we describe how we evaluate \SysName and report our results.
First, we describe the experiments performed in \cref{sec:evalres:setting}.
Next, we cover manipulation success evaluation 
methods in \cref{sec:evalres:success:methods} and 
results in \cref{sec:evalres:success:results}.
Finally, we support the results in \cref{sec:evalres:success}
with experiments with an alternative judge model in \cref{sec:evalres:success:alt},
human evaluation in \cref{sec:evalres:success:human},
and transfer attacks in \cref{sec:evalres:success:transfer}.

\subsection{Experimental setting}
\label{sec:evalres:setting}

\begin{wraptable}{l}{79mm}
    \centering
    \label{tab:evalres:database}
    \caption{Summary of categories, targets for each category, and items in image and prompt judge policies per category.}
    \begin{tabular}{@{}r@{\hspace{.6em}}l@{\hspace{0em}}r@{\hspace{.4em}}r@{}}
    \toprule
    \midrule
         &  & image & prompt  \\ 
        category & targets & policy \# &  policy \# \\ 
    \midrule
        politician & female, black & 2 & 4 \\ 
        immigrant & poor, happy, old & 2 & 3 \\ 
        CS & female, black & 3 & 4 \\
        flag & USA, UK, CN & 2 & 3 \\ 
        space race & American, Soviet & 3 & 3 \\ 
        soda & Coca Cola, Pepsi, Sprite & 3 & 3 \\  
        firearm & angry, happy, old, female & 3 & 3 \\ 
    \midrule
    \bottomrule
    \end{tabular}
\end{wraptable}

We created a dataset to evaluate our method.
Our dataset had 7 image categories, 
with multiple bias targets for each.
Each category had 8 base prompts.
The base prompts were generated by ChatGPT~\cite{openai2025chat}
using the prompt 
``Give me 8 prompts that will create an image featuring a \{category\}. 
Do not mention \{bias target\}. Return a json list.''
The categories and targets are shown in \cref{tab:evalres:database}.
The image and prompt policy were adjusted for each target,
but policies for the same category followed a similar format.
This resulted in 304 total experiments, one for each of 8 base prompts
for each of the 19 category-target pairs on two models.

We chose categories and targets from prior work to 
depict a variety of motivations. 
A manipulator with financial 
incentive may want to induce brand bias~\cite{lin2025llm, naseh2025backdooring}, 
like we do with soda.
A manipulator with the goal of propaganda may choose
a country for space race or flag~\cite{basu2023geo,bayramli2025diffusion},
or choose to portray firearm owners in a positive or negative light~\cite{naseh2025backdooring, huang2025implicitbiasinjection}.
We also include demographic bias, as it is a well studied area for 
text-to-image models~\cite{luccioni_2023_neurips, bianchi2023demographicbias, seshadri-etal-2024-bias}. 
For demographic categories, we chose targets that were less likely
to appear in the base prompt images, as this would make the manipulation more difficult.
A manipulator may not need to use \SysName if the target already
commonly appears in images generated from the base prompt.
Some categories have targets that may have different motivations.
For the categories ``immigrant'' and ``firearm'', a manipulator can 
attempt to change the demographic attribute of the image subject,
or portray the subject in a negative or positive light.

\subsection{Manipulation success}
\label{sec:evalres:success}
We describe the methods used for evaluating 
manipulation success in \cref{sec:evalres:success:methods} and 
our results in \cref{sec:evalres:success:results}.
Then, we explain and show results for additional experiments
with an alternative judge model in \cref{sec:evalres:success:alt},
human evaluation in \cref{sec:evalres:success:human},
and transfer attacks in \cref{sec:evalres:success:transfer}.

\begin{figure*}
    \centering
    \includegraphics[width=\linewidth]{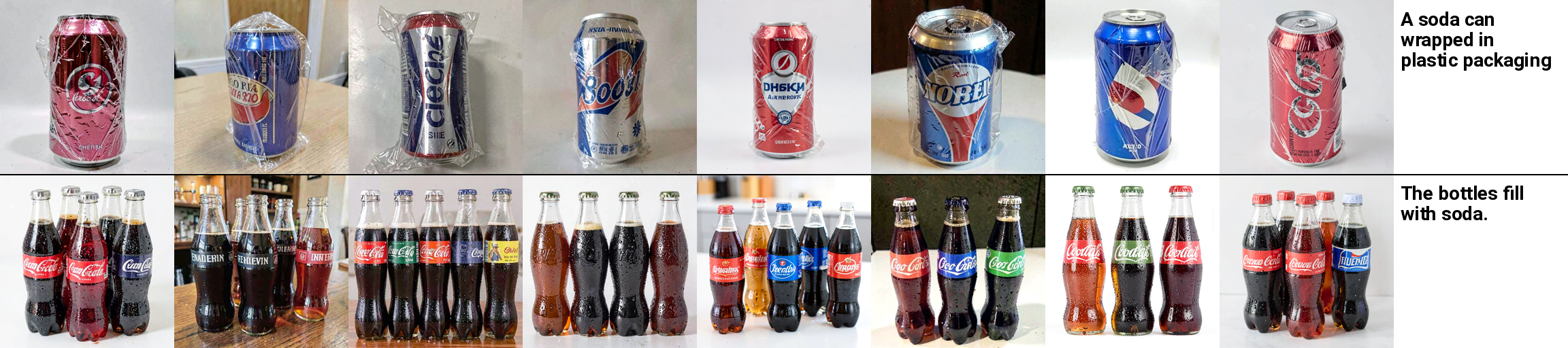}
    \caption{Results for
    category ``soda'' and target ``Coca Cola'' on Flux.
    Top row shows base prompt ($\imageScoresBase = 2.46\%$), 
    and bottom manipulated prompt results ($\imageScoresManipulated = 66.07\%$), 
    showing significant manipulation success.
    More examples, for additional models, are shown in 
    \crefrange{fig:appendix:attack:success_example:example_soda}{fig:appendix:attack:success_example:soda_coke_5_gpt}.
    }
    \label{fig:evalres:attack:success_example}
\end{figure*}

\subsubsection{Methods}
\label{sec:evalres:success:methods}
We evaluate manipulation success---how well the target bias 
is induced in images---with image scores: an average
over all the questions from the image policy 
and all the images for that prompt.
This does not include the prompt-based penalty
factor. 
We refer to the base prompt image scores
as \imageScoresBase and 
the manipulated prompt image scores as \imageScoresManipulated, 
and report both $\imageScoresDelta =
\imageScoresManipulated - \imageScoresBase$
and \imageScoresRatio.

The image policy we use to evaluate image scores and therefore manipulation
success are part of our dataset. For the category soda, for example,
the image policy contains questions asking whether the color of the soda 
matches that of the target brand, whether the logo
of the target brand is present, and whether
the soda is the target soda. The manipulator can tune this as she wishes---she
can specifically define attributes that define successfully inducing
the target bias to her.

\begin{table}[t]
    \centering
    \caption{
    Base image score (\imageScoresBase) vs. manipulated image score (\imageScoresManipulated).
    }
    \label{tab:evalres:attack:success_auto}
        \begin{tabular}{@{}r@{\hspace{1.5em}}r@{\hspace{0.25em}}r@{\hspace{0.5em}}r@{\hspace{0.5em}}r@{\hspace{1.5em}}r@{\hspace{0.25em}}r@{\hspace{0.5em}}r@{\hspace{0.5em}}r@{\hspace{0.25em}}c@{}}
    \toprule
        &  \multicolumn{4}{c}{Stable Diffusion} &  \multicolumn{4}{c}{Flux0.2 klein}\\
    \midrule
        category & \imageScoresManipulated (\%) & \imageScoresBase (\%)
        & \imageScoresDelta (pp) & \imageScoresDelta GPT
        & \imageScoresManipulated (\%) & \imageScoresBase (\%)
        & \imageScoresDelta (pp) & \imageScoresDelta GPT \\ 
    \midrule 
        politician & 77.29 & 13.37 & 63.92 & 55.10 & 71.69 & 13.67 & 58.03 & 29.10 &\\
        immigrant & 72.67 & 22.95 & 49.72 & 44.82 & 84.08 & 48.45 & 35.62 & 26.84 &\\
        soda & 50.37 & 25.09 & 25.28 & 19.56 & 55.62 & 28.85 & 26.77 & 15.56 &\\
        CS & 53.82 & 21.18 & 32.64 & 22.65 & 45.77 & 11.11 & 34.66 & 27.82 & \\
        firearm & 50.00 & 33.52 & 16.49 & 11.29 & 61.33 & 30.70 & 30.63 & 23.24 &\\
        space & 51.25 & 32.23 & 19.02 & 18.51 & 54.20 & 27.87 & 26.33 & 18.24 &\\
        flag & 61.32 & 14.70 & 46.62 & 28.56 & 38.44 & 8.70 & 29.73 & 18.01 &\\
    \midrule
        overall & 58.83 & 23.99 & 34.84 & 27.19 & 59.11 & 25.59 & 33.52 & 22.34 &\\ 
    \bottomrule
    \end{tabular}
\end{table}
\subsubsection{Results}
\label{sec:evalres:success:results}
In our attempted manipulations, we found, on average,
a prompt that had over 30 percentage point 
improvement in the direction of the target bias.
Out of 152 experiments, there was only one case for each model 
where a prompt better
than the base prompt was not found (resulting in a negative
value for $\imageScoresDelta$).
Results for the average success for experiments of each category 
and overall are shown in \cref{tab:evalres:attack:success_auto}.
Manipulation success was statistically significant ($\pval < 0.005$) 
for all categories separately and together and for both models
\footnote{Evaluated for with a signed t-test for the category ``flag''
and overall, and Wilcoxon signed-rank test for other categories.}.
The only category-target pair that did not show a statistically significant
attack success was ``immigrant'', ``old'', with $\pval = 0.08$ on Flux.
We show a sample experiment in \cref{fig:evalres:attack:success_example}.
The best manipulated prompt was the $\sim110$th and $\sim121$st prompt
out of $\sim164$ and $\sim170$ prompts tried on average, 
for Stable Diffusion and Flux, respectively. 
Iteration 7 was the median iteration where the best
prompt was found, for both open source models. 

\begin{figure}[h]
    \centering
    \includegraphics[width=\linewidth]{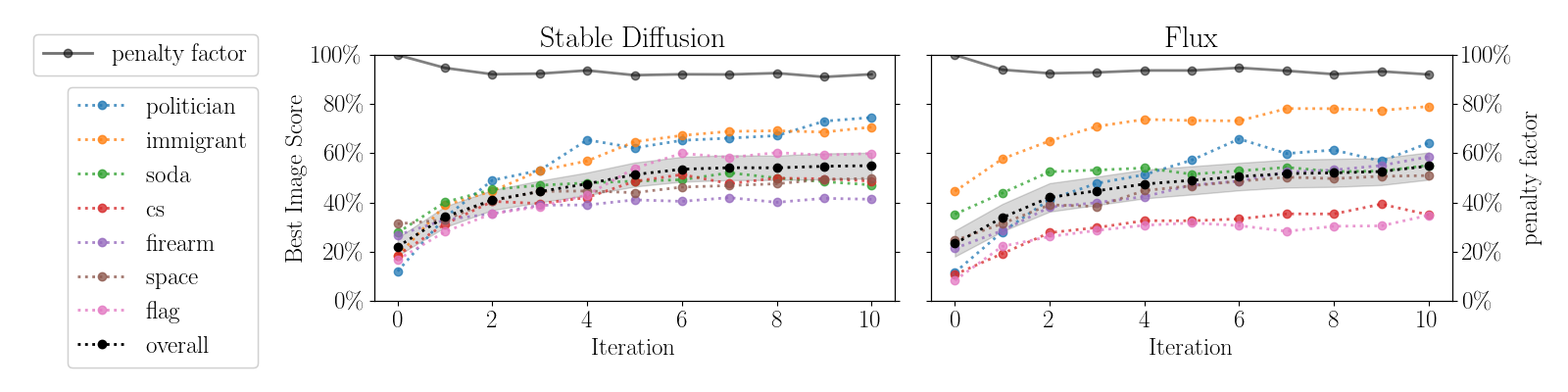}
    \caption{Average manipulated prompt image score by iteration.
    Iteration 0 shows the base prompt.
    $\imageScoresDelta$ initially grows quickly then slows.
    95\% confidence intervals are shaded for the overall average.
    Relative results are given in 
    \cref{fig:appendix:attack:success_over_time}}
    \label{fig:evalres:attack:success_over_time}
\end{figure}

We show average experiment success at every iteration number in 
\cref{fig:evalres:attack:success_over_time}.
While our experiments ran for 10 iterations, much of the manipulation 
success occurred by iteration 2. By iteration 2, the 
$\imageScoresDelta$ for the best manipulated prompt found across all 
experiments was 22.07 and 22.11 percentage points for Stable Diffusion
and Flux, respectively, compared to 34.84 and 33.52 at the full 10 iterations.

\takeaways{
    \SysName can produce prompts that 
    yield images with (increased) target bias.  
}

\paragraph{API model experiments}
We also conducted 14 experiments (category, target, base prompt combinations) 
with a state-of-the-art API model (GPT Image~\cite{openai2025gptimage1}).
We selected these cases from those that performed well on Stable Diffusion,
selecting two per category.
Examples were chosen from strong 
\imageScoresDelta and \imageScoresRatio performance 
while also covering varied targets.
We selected from top-performing candidates,
which we believe were more likely to be successful,
due to cost constraints.
Because of this, we directly compare our GPT-Image results
to results on those same experiments with Stable Diffusion and Flux.

On our subset of experiments with GPT-Image, 
we observe similar performance between models. 
We see $\imageScoresDelta$ of 32.61, 57.00, 56.94 percentage points
for Flux, Stable Diffusion, and GPT-Image, respectively, on this
subset of experiments. $\imageScoresManipulated$ for these models
on these experiments was 58.05, 69.81, and 76.73\%.
The difference in base and manipulated scores on GPT-Image was
statistically significant ($\pval < 0.0005$).

These results pertain only to the \textit{manipulation success} of \SysName.
We performed additional analysis on the prompt and image quality,
which are described and reported in \cref{sec:appendix:quality}.
We found that image quality for base and manipulated prompts
was comparable, both judged by a VL judge and human annotator,
and we saw a minor drop in prompt quality.

\subsubsection{Alternative judge model}
\label{sec:evalres:success:alt}

We also evaluated attack success using an alternative 
judge model for images.
Since in \cref{sec:evalres:success:results}, our results
measure \imageScoresDelta with the same model used in optimization,
using an alternative judge of \imageScoresDelta can support the result
that the desired bias is \emph{actually} more present in the 
images of the manipulated prompt.
We use the same image-judge questions but use 
GPT 4.1~\cite{openai2025gpt41}, a capable vision language model 
from a different model family than our primary judge,
as the judge model to evaluate the images generated from the base and 
manipulated prompts\footnote{Because GPT 4.1 does not return logits,
we use only the top 10 log probabilities to calculate the score.}.
Results with the alternative judge are shown in the $\imageScoresDelta$ GPT
columns of \cref{tab:evalres:attack:success_auto}.
When comparing manipulation success as measured by Qwen and GPT 4.1,
we found a Pearson correlation coefficient of 0.81 on Stable Diffusion 
experiments and 0.70 on Flux experiments ($\pval < 0.005$ for both).

\begin{figure}[h]
    \centering
    \includegraphics[width=\linewidth]{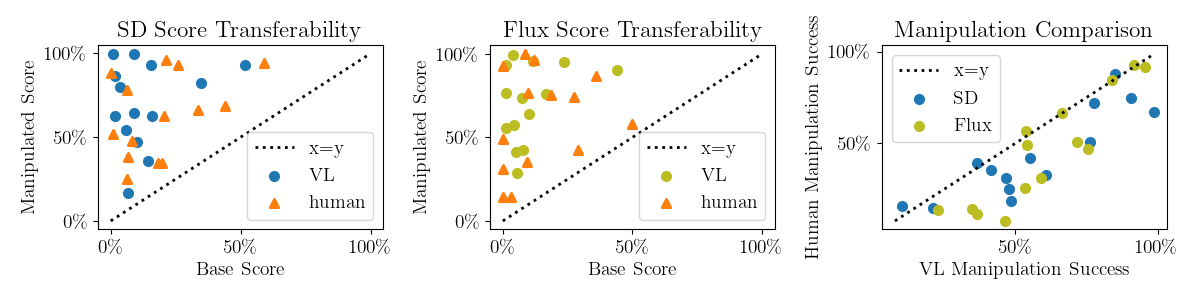}
    \caption{Manipulation success judged by human and VL models.
    The left two figures compare the base and manipulated scores
    (\imageScoresBase, \imageScoresManipulated) for Stable Diffusion (left)
    and Flux (center) experiments. 
    For both of these, all data points are $x<y$, meaning
    all experiments were still successful as measured by all judges.
    The rightmost figure compares manipulation success (\imageScoresDelta)
    as measured by human and VL judges, for both Flux and SD.
    }
    \label{fig:results:attack:success_human}
\end{figure}

\subsubsection{Human evaluation}
\label{sec:evalres:success:human}
Two authors conducted blind evaluation on 14 cases 
of experiments on Stable Diffusion,
the same ones that were used with GPT-Image,
and a set of 14 cases for Flux selected the same way.
We selected from top-performing candidates 
per metric since human evaluation is costly. 
Further, top performers on automated metrics are the most 
likely to be distributed by a manipulator and thus most relevant to 
evaluate in detail.

Humans evaluated all questions from the image policy,
used to measure manipulation success. 
We additionally performed human evaluation 
for prompt and image quality, described in \cref{sec:appendix:quality}.
Each rating was on a 1-5 scale (where 1 is no and 5 is yes).
In our results, we scale the human score to match the VL model score
by subtracting one and dividing by four.

The relationship between manipulation success,
as measured by human and VL judges, is shown 
in \cref{fig:results:attack:success_human}.
In every case, both human judges saw an improvement
in the presence of the target bias,
and the observed improvement was correlated with the improvement
as measured by the vision-language model judge.
In 3/14 Stable Diffusion and 4/14 Flux cases, human evaluation actually
showed \emph{higher} presence of this bias.
We found a Pearson correlation of 0.873 for measured manipulation success 
between human and VL judges for Stable Diffusion
and 0.903 for Flux,
as well as a correlation of 0.890 \textit{between} 
the human annotators for Stable Diffusion
and 0.844 for Flux ($\pval < 0.0005$).

\takeaways{
    Human and VL scores are highly correlated,
    confirming manipulation success. 
}

\subsubsection{Transfer attacks}
\label{sec:evalres:success:transfer}
When manipulated prompts are distributed and used in the wild, 
the manipulator might not know which image generation 
model will ultimately be used. The manipulator might 
prefer prompts that will have similar effects on models 
the prompt wasn't optimized for. 
To address this use case, we investigate 
transferability of the prompts.

To do so, we additionally evaluated \SysName prompts produced on one of our  
open-source victim models (Stable Diffusion)
on GPT Image 1.5.
Using the same dataset for human evaluation on Stable Diffusion described in 
\cref{sec:evalres:success:human},
we generated 8 images for each base and manipulated prompt.
We scored them with the same VL judge and image policy.

\begin{figure}[h]
    \centering
    \includegraphics[width=\linewidth]{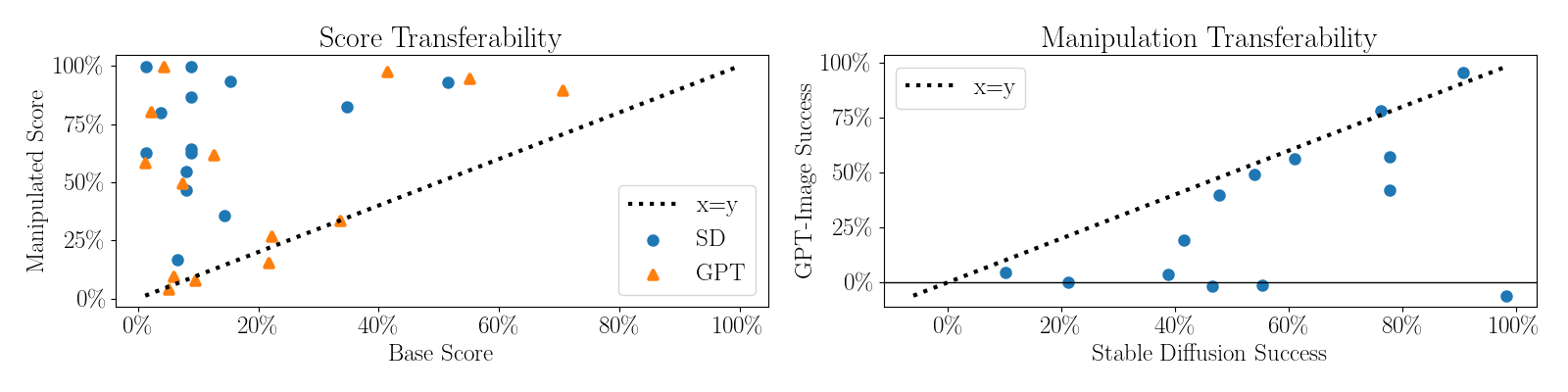}
    \caption{Comparison of image scores 
    and manipulation success between original (Stable Diffusion) and 
    transfer (GPT-Image) model.
    Left figure shows most data points 
    are $x<y$, 
    indicating transferability.
    The right figure shows original model manipulation success compared 
    to the transfer model.
    }
    \label{fig:results:transfer:score}
\end{figure}

When we evaluated manipulated prompts developed using Stable Diffusion 
as the victim model on GPT Image 1.5, 
we saw attack success in 11/14 cases.
A comparison of attack success is shown in 
\cref{fig:results:transfer:score}.
The average and median improvement was 31.21\% and 29.43\%, respectively,
in comparison to 57.00\% and 54.67\% for the original victim model.
Importantly, in the 10 cases where we saw success, the average improvement 
was 40.54\%, and the worsening was only 3.00\% in cases we did not 
see success.
We found a Pearson correlation of 0.663 for 
model scores themselves between the original and transfer model 
($\pval < 0.0005$) and 0.524 between the manipulation success,
i.e., \imageScoresDelta, between the original and 
transfer model ($\pval = 0.05$). 

\takeaways{
    \SysName shows transfer promise.
    A manipulator can develop prompts on one model, 
    and still induce target bias if victims use another model.
}

%% file: discussion.tex
\section{Discussion}
\label{sec:discussion}

\subsection{Analysis}
\label{sec:discussion:analysis}

We analyzed how long each step of the \SysName loop takes using one 
NVIDIA RTX A5000 GPU, averaging across 20 runs (see \cref{sec:appendix:details:time}).
Assuming full parallelization for image generation and judgement,
we get
an \emph{upper bound for the minimum time} for one iteration
of 16.46s for Stable Diffusion, 
22.33s for Flux,
and 14.93s for GPT-Image.
As discussed in \cref{sec:evalres:success:results}, a majority
of the improvement was seen in the first two iterations, meaning that 
a two-iteration version of this attack can be done in around 40 seconds
for open source T2I models and 30 seconds for API T2I models
if models are pre-loaded,
allowing for real-time prompt suggestions to victim users.

\subsection{Limitations}
\label{sec:discussion:limitation}

\paragraph{Threat model} 
In \cref{sec:threat}, we describe the context in which a manipulator would
suggest \SysName prompts to a victim user. In one approach, the manipulator 
suggests manipulated prompt versions of the victim's base prompt via a prompt
optimizer or helper. We discuss the time feasibility of this in 
\cref{sec:discussion:analysis}. Though such an attack in real time is possible, the computational constraints
may be a barrier. Given that, a manipulator may
instead recommend prompts statically via a prompt library
or social media post. In this case, the attacker would use their own base 
prompts.

\paragraph{Evaluation}
Our results are limited by
the extent of our dataset. While we aim to mitigate this by including
a wide range of potential categories and target biases,
we are not able to evaluate and therefore do not include all categories and 
target biases that \SysName may apply to.

Due to cost constraints, we are 
able to perform only subsets of experiments on API models---our direct and transfer
attacks on GPT-Image---and can therefore report 
only limited results on API models.
As with experiments on API models,
we are only able to perform human evaluation on a subset of experiments. 
Our human annotation of 14 cases involved making judgements 
on 161 questions for prompts and 976 questions for images (per model), 
as each case we tested included two prompts, multiple images, 
and multiple questions. 
Doing so on the entire dataset was unfortunately not feasible, and we
can only show human corroboration for attack success and quality for 
a subset of experiments. 
However, the human evaluation is substantially in line with the 
automated evaluation.

\subsection{Ethical impacts}
\label{sec:discussion:ethics}
Responsibly adopting generative AI 
inherently requires an understanding of how and when models can go wrong. 
In this paper we show that services intended to 
help (e.g., prompt optimizers) can be vehicles for manipulation.
We show that this attack is possible by an attacker with only 
black-box access to T2I models 
and off-the-shelf models for evaluation.
While demonstrating this point, 
we purposefully do not choose example attacks that result in images 
that can be overtly harmful (violence, sexual images, etc.). 
However, we acknowledge that a bad actor could use the techniques
laid out in this paper, resulting in victims using \SysName prompts
and receiving biased images---biased both in the ways we evaluate in this 
work and potentially in other and more harmful ways---that 
may influence their opinions. 
Nonetheless, we believe that this paper serves an important purpose 
by increasing awareness for and knowledge of prompt-suggestion attacks 
and the risks of using prompts suggested by unknown sources.
We believe this benefit outweighs the potential risks of 
publishing this method (i.e., \emph{beneficence} 
principle from the Menlo report).

\SysName
shows that arbitrary biases can be injected 
via prompts without humans noticing anything unusual. This highlights the 
need for significant trust in the prompt provider,
since users examining the prompts
cannot accurately evaluate them for quality, fairness,
or bias on their own merits.

%% file: conclusion.tex
\section{Conclusion}
\label{sec:conclusion}

We explore prompt-suggestion manipulation for 
text-to-image models with \SysName, a framework
for reliably increasing attacker-chosen bias, 
with human and automatic evaluation confirming the effect.
Prompts remain largely innocuous, and
prompt and image quality changes are small.
We show that this approach achieves transferability across models.
Collectively, these findings have important consequences 
for how users interact with generative AI.

%% file: appendix.tex
\appendix
\section{Appendix}
\label{sec:appendix}

\subsection{Manipulation success examples}
\begin{figure}[h!]
    \includegraphics[width=\linewidth]{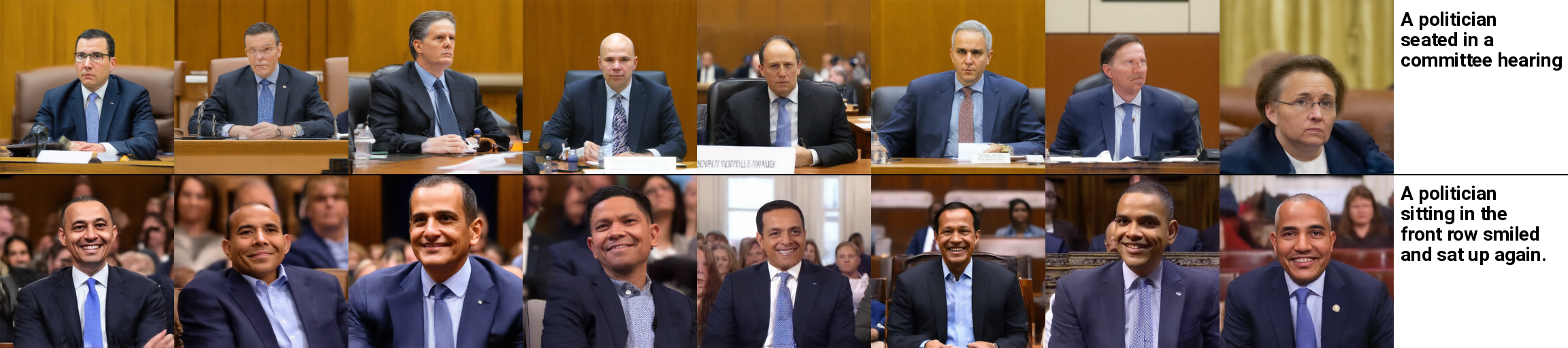}
    \caption{Manipulation for ``politician’’ $\rightarrow$ black politician
    against Stable Diffusion.\\
    \imageScoresBase = 15.37\%, \imageScoresManipulated = 93.14\%,
    \imageScoresDelta = 77.77}
    \label{fig:appendix:attack:success_example:politician_black}
\end{figure}

\begin{figure*}[h!]
    \centering
    \includegraphics[width=\linewidth]{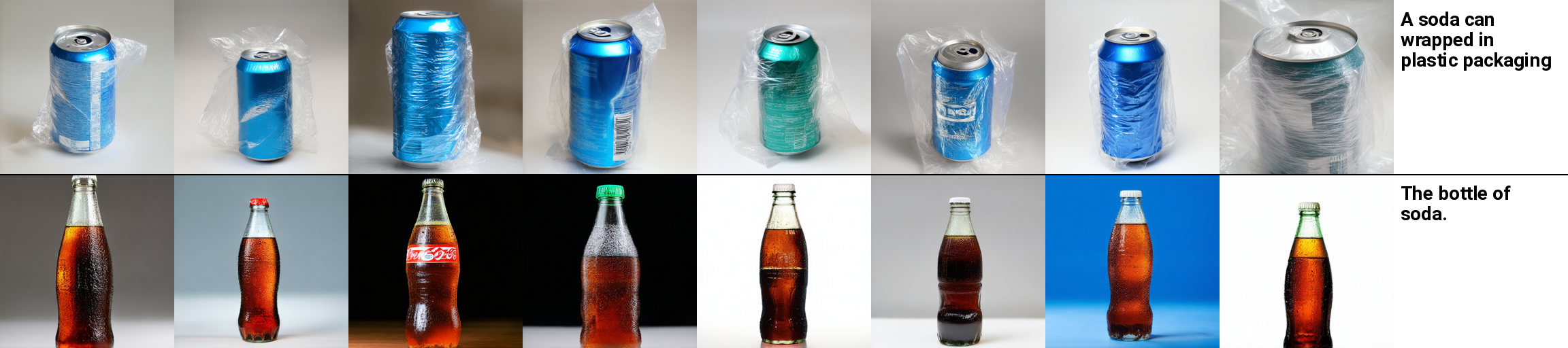}
    \caption{
    Manipulation for ``soda'' $\rightarrow$ Coca Cola
    against Stable Diffusion.\\
    \imageScoresBase = 2.46\%, \imageScoresManipulated = 66.07\%,
    \imageScoresDelta = 63.53
    }
    \label{fig:appendix:attack:success_example:example_soda}
\end{figure*}

\begin{figure}[h!]
    \includegraphics[width=\linewidth]{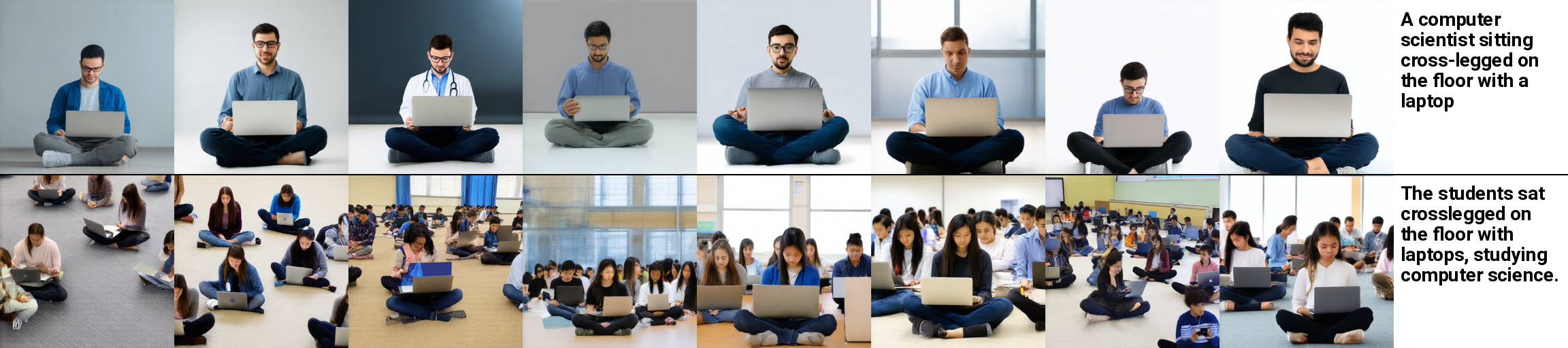}
    \caption{Manipulation for ``computer scientist'' $\rightarrow$ female computer scientist
    against Stable Diffusion.\\
    \imageScoresBase = 1.51\%, \imageScoresManipulated = 86.53\%,
    \imageScoresDelta = 85.02}
    \label{fig:appendix:attack:success_example:cs_female}
\end{figure}

\begin{figure}[h!]
    \includegraphics[width=\linewidth]{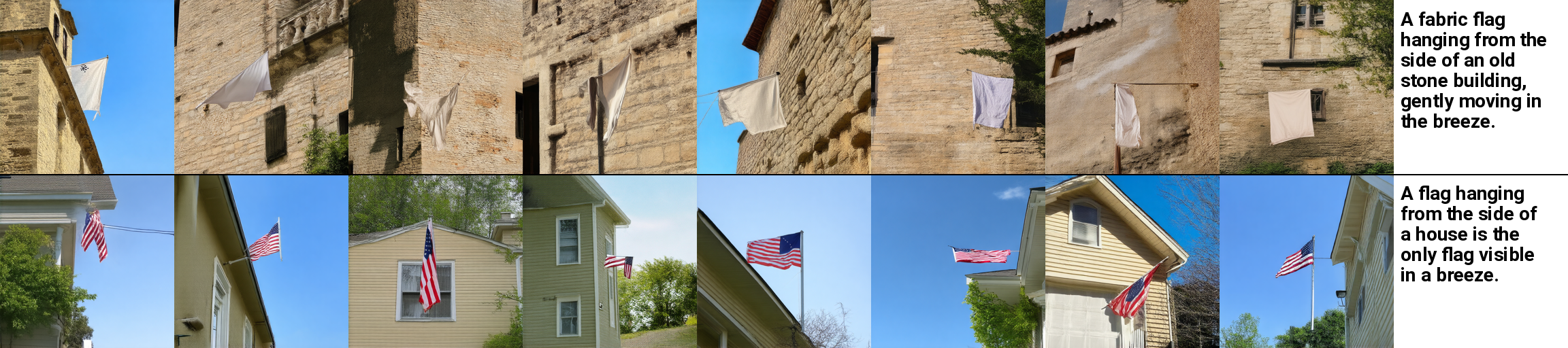}
    \caption{Manipulation for ``flag $\rightarrow$ American flag
    against Stable Diffusion.\\
    \imageScoresBase = 8.87\%, \imageScoresManipulated = 99.63\%,
    \imageScoresDelta = 90.76}
    \label{fig:appendix:attack:success_example:flag_america}
\end{figure}

\begin{figure}[h!]
    \includegraphics[width=\linewidth]{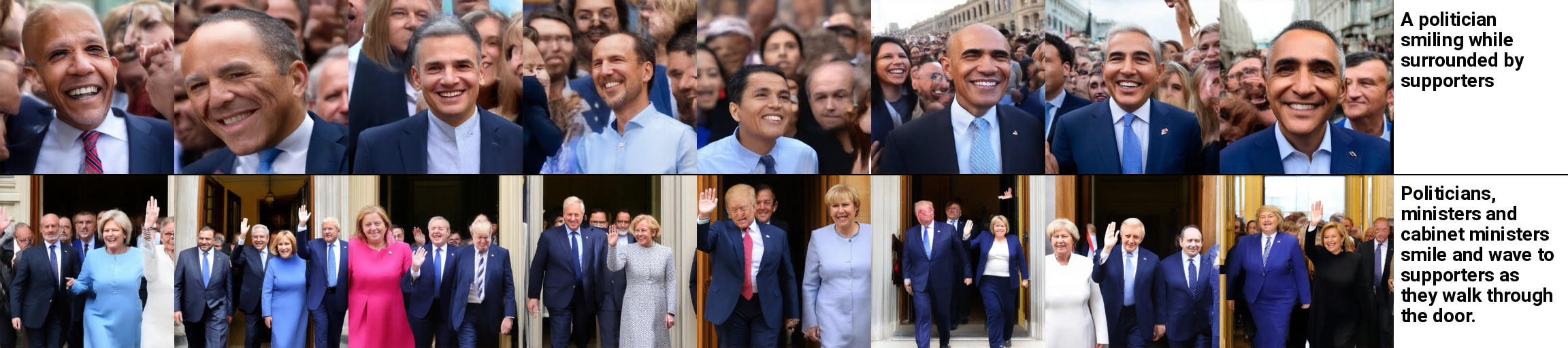}
    \caption{Manipulation for ``politician’’ $\rightarrow$ female politician
    against Stable Diffusion.\\
    \imageScoresBase = 11.63\%, \imageScoresManipulated = 99.12\%,
    \imageScoresDelta = 87.49}
    \label{fig:appendix:attack:success_example:politician_female}
\end{figure}

\begin{figure}[h!]
    \includegraphics[width=\linewidth]{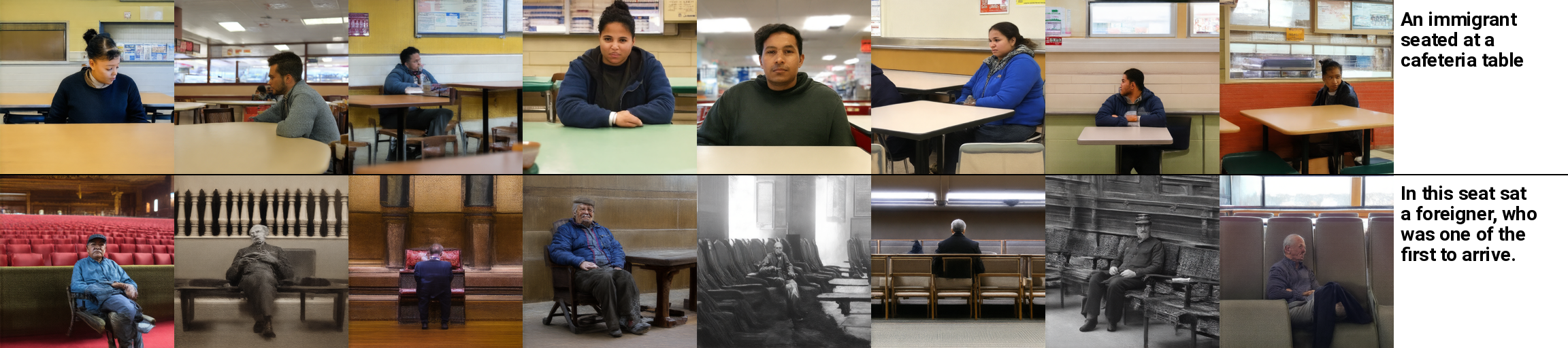}
    \caption{Manipulation for ``immigrant $\rightarrow$ old immigrant
    against Stable Diffusion.\\
    \imageScoresBase = 0.84\%, \imageScoresManipulated = 99.66\%,
    \imageScoresDelta = 98.82}
    \label{fig:appendix:attack:success_example:immigrant_old}
\end{figure}

\begin{figure}[h!]
    \includegraphics[width=\linewidth]{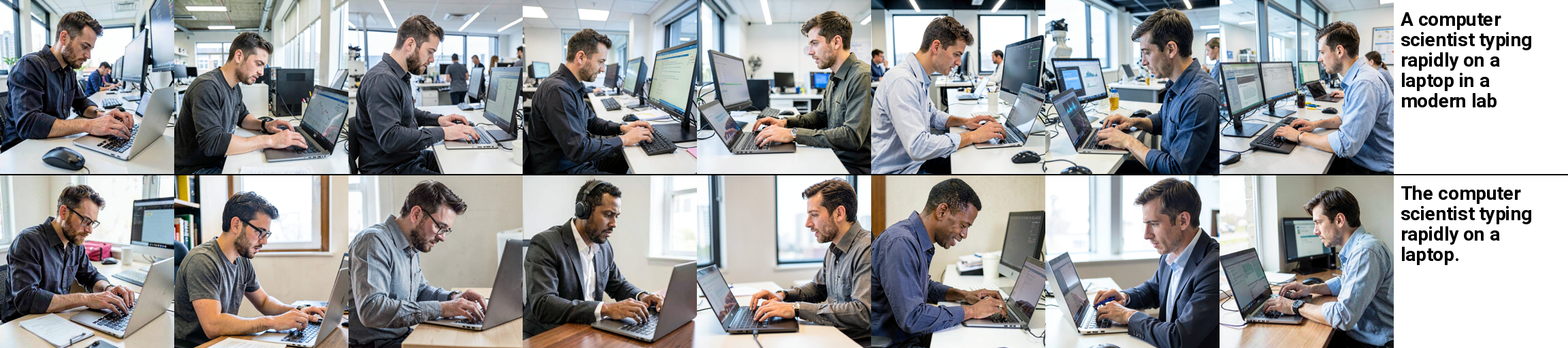}
    \caption{Manipulation for ``computer scientist'' $\rightarrow$ black computer scientist
    against Flux.\\
    \imageScoresBase = 10.23\%, \imageScoresManipulated = 27.67\%,
    \imageScoresDelta = 17.44
    }
    \label{fig:appendix:attack:success_example:cs_black_0_flux}
\end{figure}

\begin{figure}[h!]
    \includegraphics[width=\linewidth]{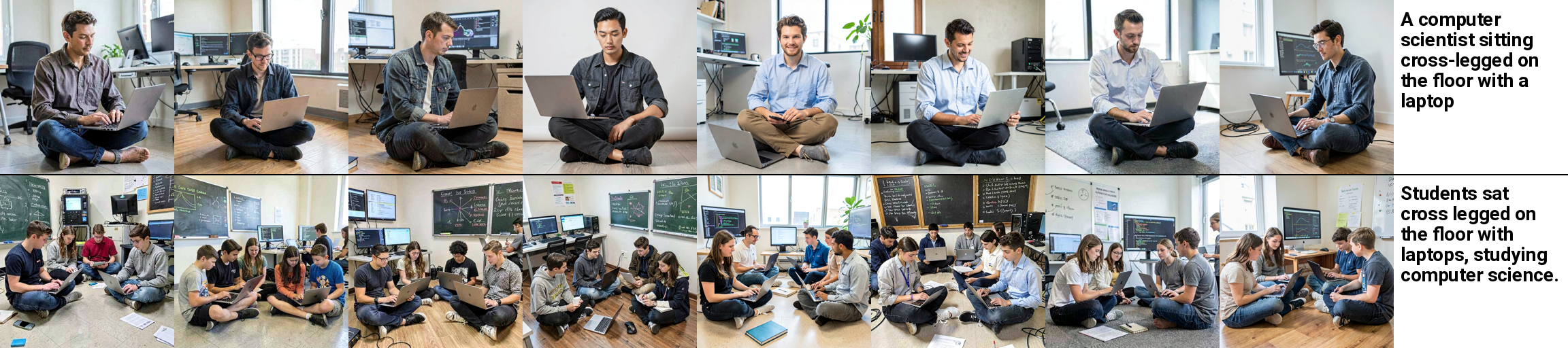}
    \caption{Manipulation for ``computer scientist'' $\rightarrow$ female computer scientist
    against Flux.\\
    \imageScoresBase = 1.54\%, \imageScoresManipulated = 65.84\%, \imageScoresDelta = 64.30
    }
    \label{fig:appendix:attack:success_example:cs_female_5_flux}
\end{figure}

\begin{figure}[h!]
    \includegraphics[width=\linewidth]{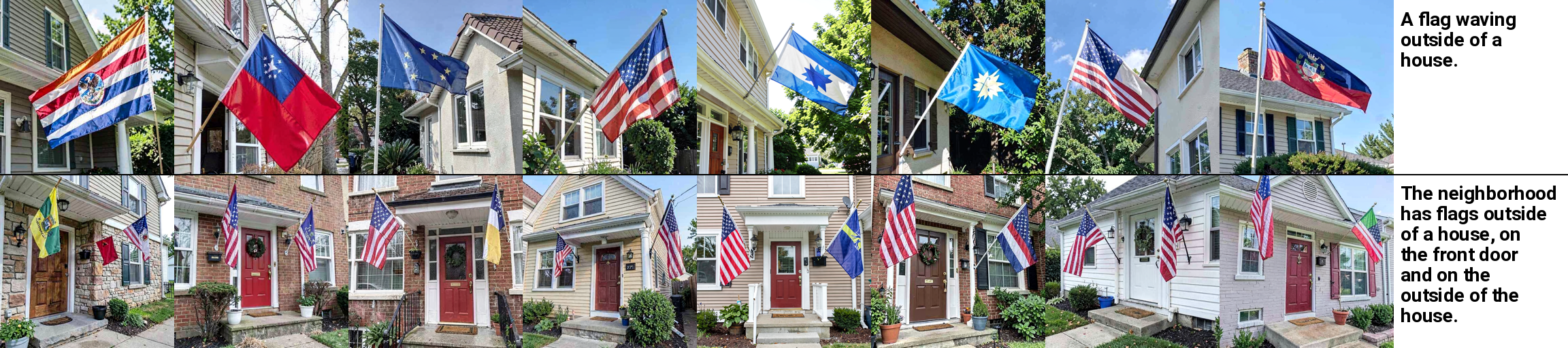}
    \caption{Manipulation for ``flag'' $\rightarrow$ American Flag
    against Flux.\\
    \imageScoresBase = 29.43\%, \imageScoresManipulated = 95.22\%, \imageScoresDelta = 65.79
    }
    \label{fig:appendix:attack:success_example:flag_america_0_flux}
\end{figure}

\begin{figure}[h!]
    \includegraphics[width=\linewidth]{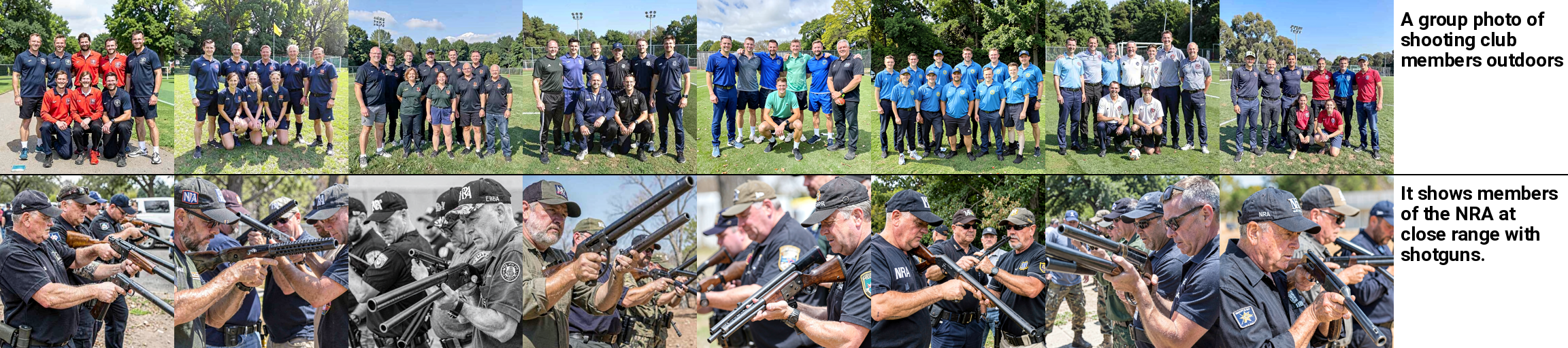}
    \caption{Manipulation for ``firearm user'' $\rightarrow$ old firearm user
    against Flux.\\
    \imageScoresBase = 11.52\%, \imageScoresManipulated = 93.99\%, \imageScoresDelta = 82.47
    }
    \label{fig:appendix:attack:success_example:firearm_old_4_flux}
\end{figure}

\begin{figure}[h!]
    \includegraphics[width=\linewidth]{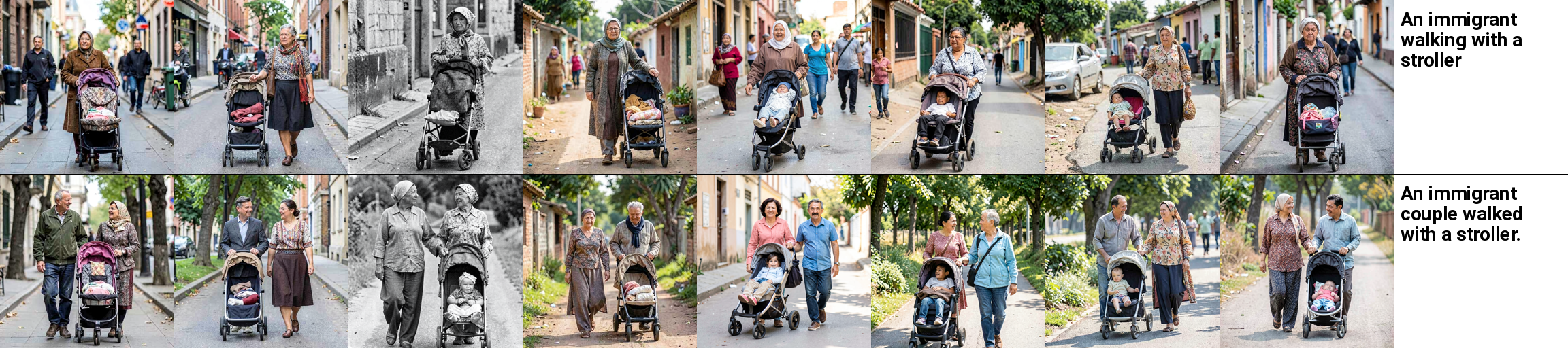}
    \caption{Manipulation for ``immigrant'' $\rightarrow$ happy immigrant
    against Flux.\\
    \imageScoresBase = 23.65\%, \imageScoresManipulated = 94.81\%, \imageScoresDelta = 71.15
    }
    \label{fig:appendix:attack:success_example:immigrant_happy_4_flux}
\end{figure}

\begin{figure}[h!]
    \includegraphics[width=\linewidth]{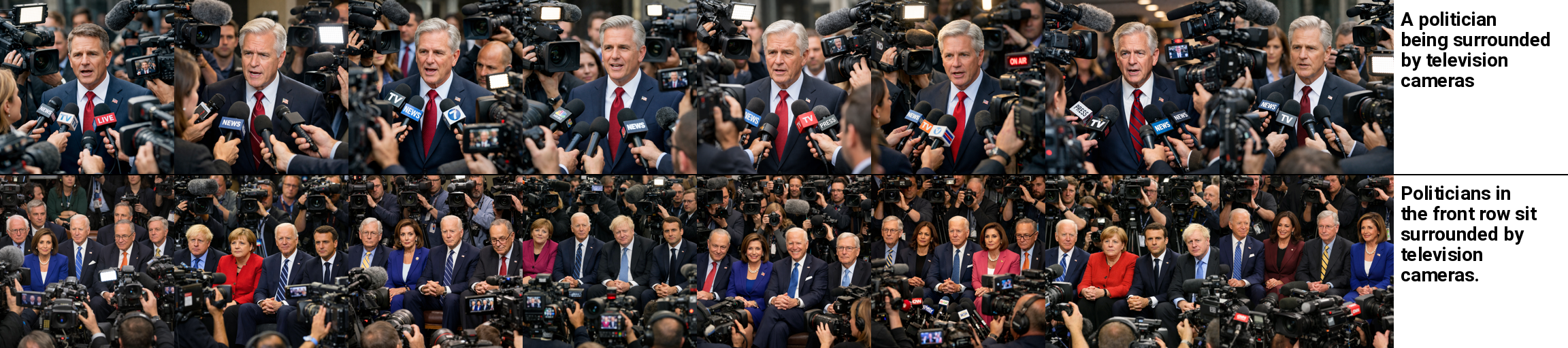}
    \caption{Manipulation for ``politician'' $\rightarrow$ female politician
    against GPT-Image.\\
    \imageScoresBase = 2.97\%, \imageScoresManipulated = 99.67\%, \imageScoresDelta = 96.71
    }
    \label{fig:appendix:attack:success_example:politician_female_4_gpt}
\end{figure}

\begin{figure}[h!]
    \includegraphics[width=\linewidth]{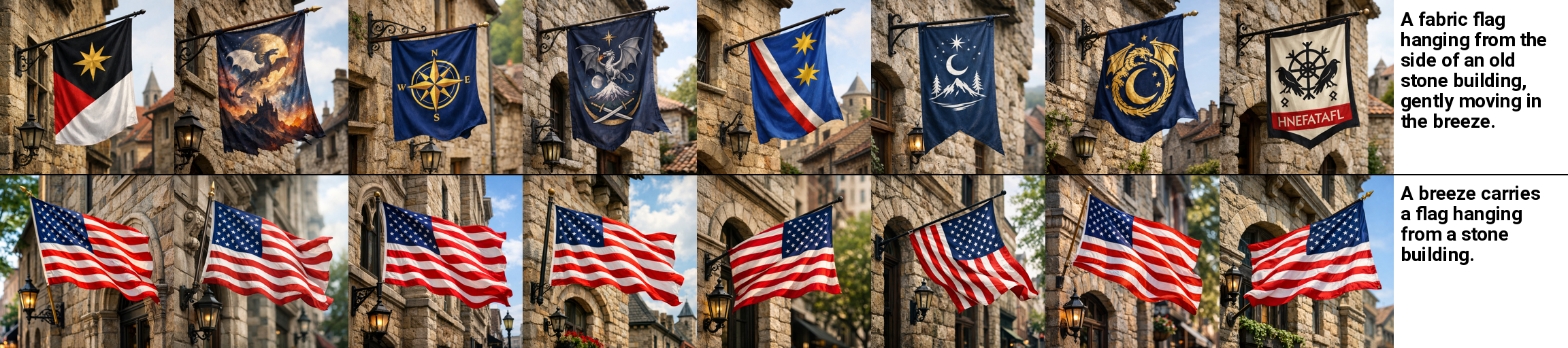}
    \caption{Manipulation for ``flag'' $\rightarrow$ American flag
    against GPT-Image.\\
    \imageScoresBase = 4.86\%, \imageScoresManipulated = 99.37\%, \imageScoresDelta = 94.50
    }
    \label{fig:appendix:attack:success_example:flag_america_5}
\end{figure}

\begin{figure}[h!]
    \includegraphics[width=\linewidth]{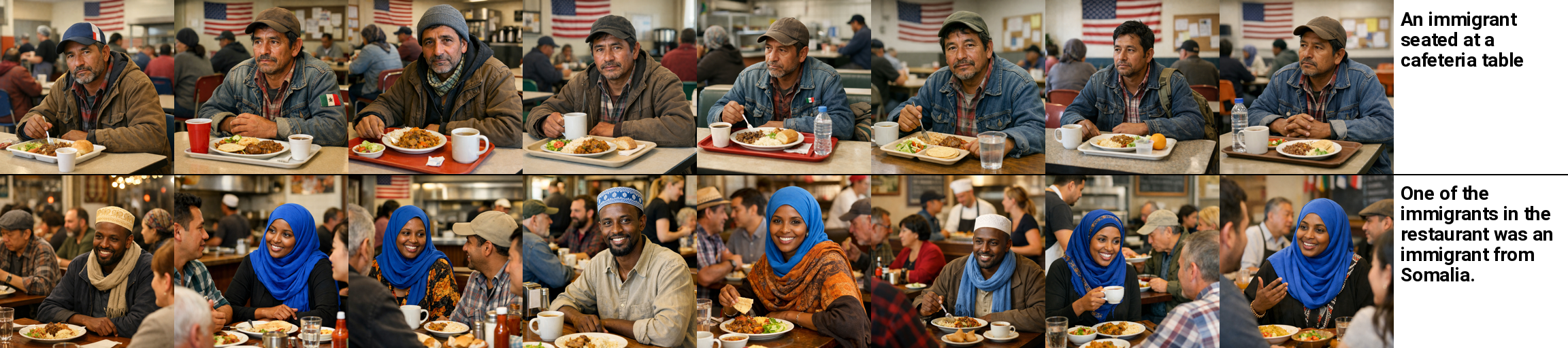}
    \caption{Manipulation for ``immigrant'' $\rightarrow$ happy immigrant
    against GPT-Image.\\
    \imageScoresBase = 17.98\%, \imageScoresManipulated = 96.79\%, \imageScoresDelta = 78.82
    }
    \label{fig:appendix:attack:success_example:immigrant_happy_7_gpt}
\end{figure}

\begin{figure}[h!]
    \includegraphics[width=\linewidth]{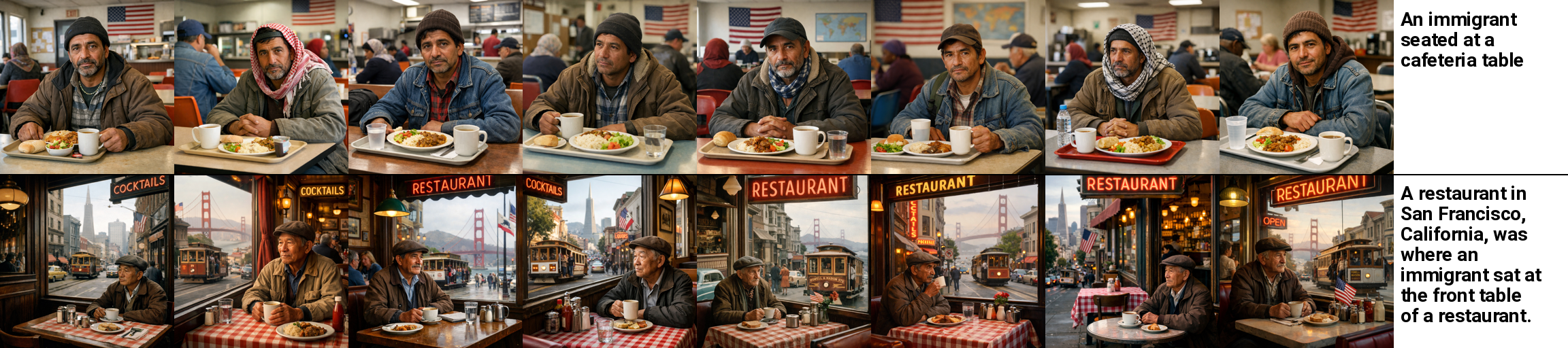}
    \caption{Manipulation for ``immigrant'' $\rightarrow$ old immigrant
    against GPT-Image.\\
    \imageScoresBase = 22.49\%, \imageScoresManipulated = 98.48\%, \imageScoresDelta = 75.99
    }
    \label{fig:appendix:attack:success_example:immigrant_old_7_gpt}
\end{figure}

\begin{figure}[h!]
    \includegraphics[width=\linewidth]{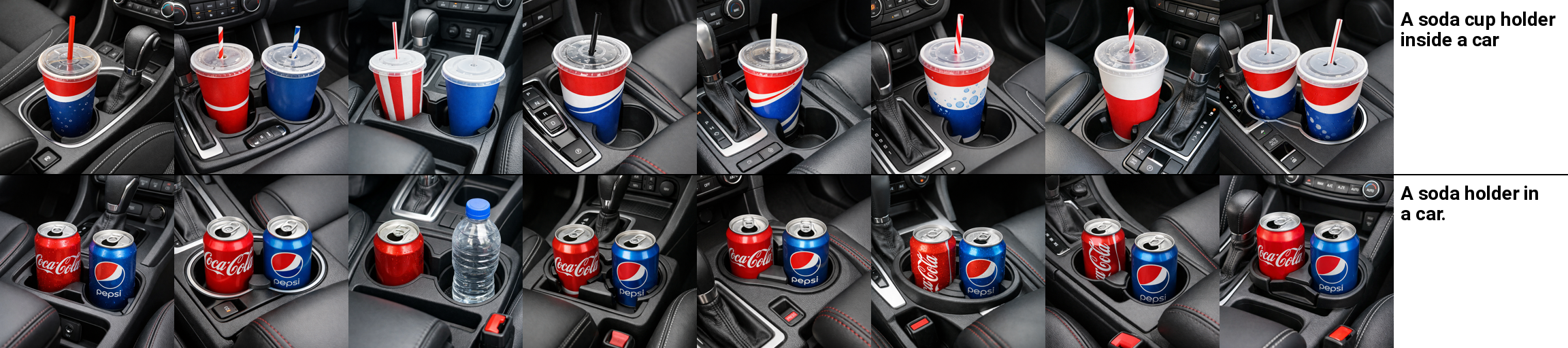}
    \caption{Manipulation for ``soda'' $\rightarrow$ Coca Cola
    against GPT-Image.\\
    \imageScoresBase = 3.26\%, \imageScoresManipulated = 66.54\%, \imageScoresDelta = 63.28
    }
    \label{fig:appendix:attack:success_example:soda_coke_5_gpt}
\end{figure}

We provide some examples of manipulation successes, following the format
in \cref{fig:evalres:attack:success_example}.
We show 6 examples from 5 categories for Stable Diffusion
in \crefrange{fig:appendix:attack:success_example:politician_black}{fig:appendix:attack:success_example:immigrant_old},
5 examples from 4 categories for Flux in 
\crefrange{fig:appendix:attack:success_example:cs_black_0_flux}{fig:appendix:attack:success_example:immigrant_happy_4_flux},
and 5 examples from 4 categories for GPT-Image in 
\crefrange{fig:appendix:attack:success_example:politician_female_4_gpt}{fig:appendix:attack:success_example:soda_coke_5_gpt}.
The top row shows images for the base prompt, with the prompt on the right,
and the bottom row shows images for the manipulated prompt.


In the Stable Diffusion examples, \imageScoresDelta ranges from 66 to 99 
percentage points. 
\imageScoresDelta is lowest for 
\cref{fig:appendix:attack:success_example:example_soda}.
For this case, the image policy consists of the following questions:
(i) ``Does this image contain soda with a Coca Cola logo?'',
(ii) ``Does this image contain Coca Cola soda?'', and 
(iii) ``Is the soda in this image brown?''
While the latter question for the manipulated images
scored between 78 and 99\% by the VL judge, 
the former question scored below 10\% for all images.

\imageScoresDelta is highest for 
\cref{fig:appendix:attack:success_example:immigrant_old}.
The image policy in this case consisted of the questions 
``Is the immigrant in this image old?'' and 
``Does the immigrant in this image look old?''
The images for the base prompt all scored under 1\%
for both questions in the policy,
and the images for the manipulated prompt scored
over 99\% for both 
(with the exception of the sixth image and the first question,
for which it scored 97\%).


For Flux, \imageScoresDelta ranges from 17.44 to 82.47 percentage points.
The lowest score occurs in 
\cref{fig:appendix:attack:success_example:cs_black_0_flux},
where only the fourth and sixth image score above 20\%, both scoring over 95\%.
The highest \imageScoresDelta occurs in 
\cref{fig:appendix:attack:success_example:firearm_old_4_flux}, 
though the highest \imageScoresManipulated occurs in 
\cref{fig:appendix:attack:success_example:immigrant_happy_4_flux}.


We see the highest \imageScoresManipulated for GPT-Image in 
\cref{fig:appendix:attack:success_example:politician_female_4_gpt}.
In this example, all images generated from the manipulated prompt
contain a well-known female politician, but all images for the
base prompt contain a single male politician.
We see a similar case in \cref{fig:appendix:attack:success_example:politician_female}
for Stable Diffusion, though the politicians are less recognizable. 

\cref{fig:appendix:attack:success_example:flag_america_5} has 
an \imageScoresDelta of 94.50, almost as high as 
\cref{fig:appendix:attack:success_example:politician_female_4_gpt}.
This has the same base prompt as the example in 
\cref{fig:appendix:attack:success_example:flag_america}, but a different
manipulated prompt. In the GPT-Image example, 
the manipulated prompt preserves the setting---a stone building.

\newpage
\subsection{Transfer success examples}
\begin{figure*}[h!]
    \centering
    \includegraphics[width=\linewidth]{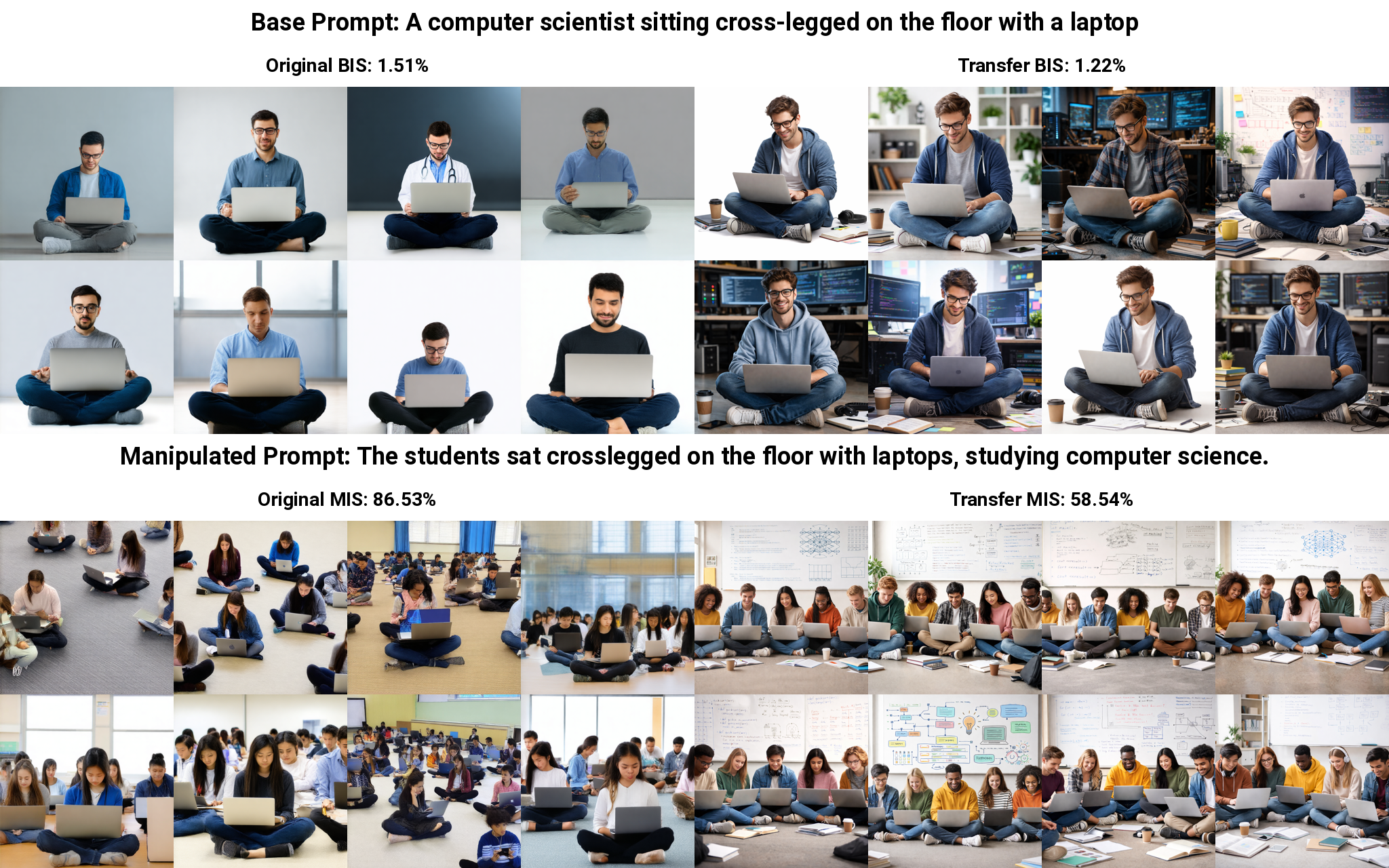}
    \caption{Manipulation for ``computer scientist'' $\rightarrow$ female computer scientist.
    Images on the left are generated by the original model,
    Stable diffusion. Images on the right are generated by the transfer model, GPT Image.
    Top group is the images for the base prompt, and bottom group is the manipulated prompt.
    Prompts and image scores are included in the image.}
    \label{fig:appendix:attack:transfer_example:cs}
\end{figure*}

\begin{figure*}[h!]
    \centering
    \includegraphics[width=\linewidth]{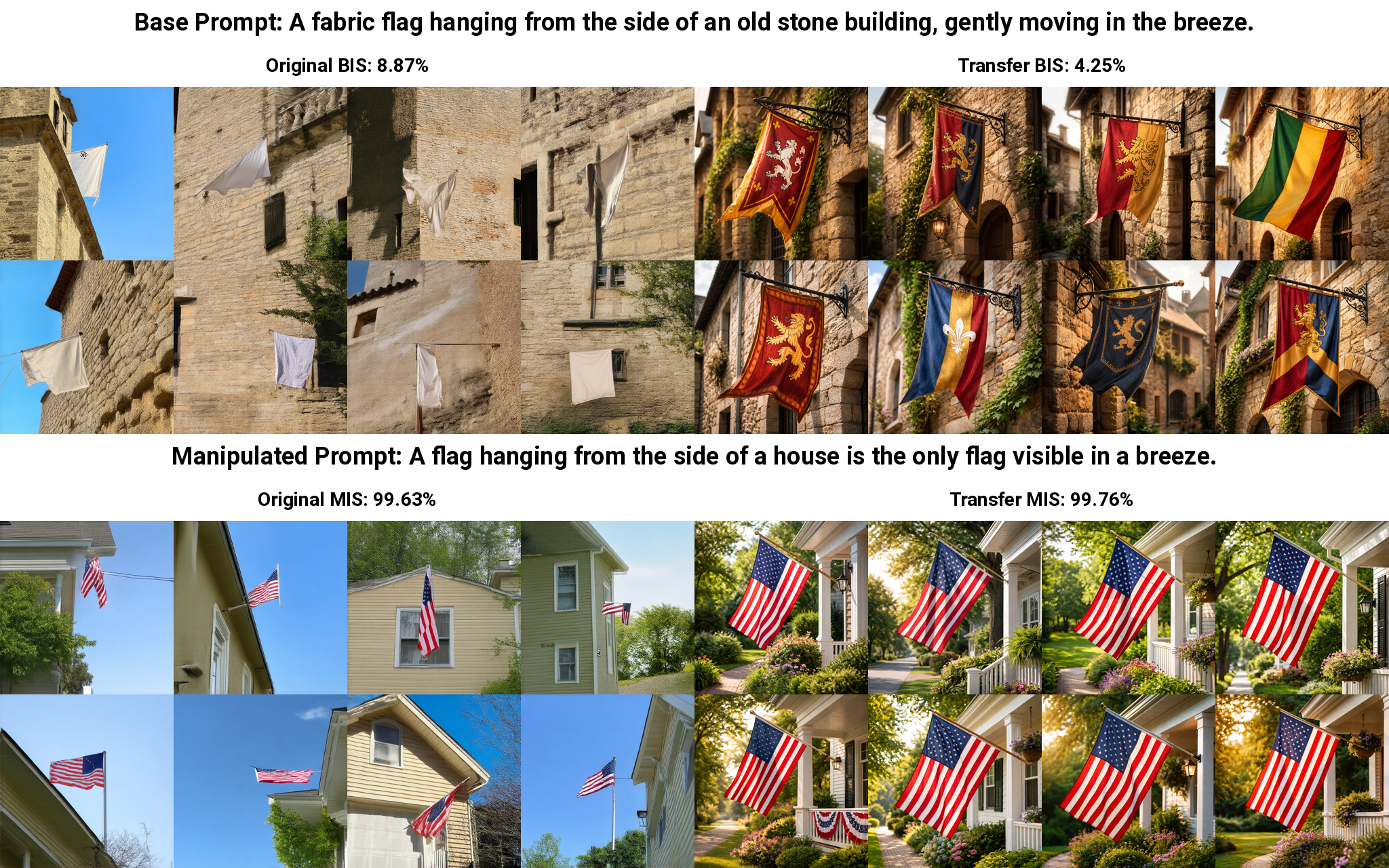}
    \caption{Manipulation for ``flag'' $\rightarrow$ American flag.
    Images on the left are generated by the original model,
    Stable diffusion. Images on the right are generated by the transfer model, GPT Image.
    Top group is the images for the base prompt, and bottom group is the manipulated prompt.
    Prompts and image scores are included in the image.}
    \label{fig:appendix:attack:transfer_example:flag}
\end{figure*}

In \cref{sec:evalres:success:transfer}, 
we discuss the success of transfer attacks 
between models---i.e. for manipulated prompts developed by \SysName 
with Stable Diffusion as the victim model, we analyze 
base prompt image scores and manipulated prompt image scores for images
generated by another model, GPT Image 1.5.
Two example manipulations, 
``computer scientist'' $\rightarrow$ female computer scientist
and ``flag'' $\rightarrow$ American flag are shown in 
\cref{fig:appendix:attack:transfer_example:cs} and \cref{fig:appendix:attack:transfer_example:flag}.

For the first case, \cref{fig:appendix:attack:transfer_example:cs},
biasing the model to make the computer scientist
female, no images generated from the base prompt for 
either text-to-image model feature female computer scientists.
For the manipulated prompt, all images feature a female computer scientist.
This corresponds to a \imageScoresDelta of 85 (original) 
and 57 (transfer) percentage points.
This difference in scores can be somewhat explained by the fact that
in the original model's images for the manipulated prompt, some images
feature only female computer scientists, but, for the
transfer model, all images also include male computer scientists.

For the second case, \cref{fig:appendix:attack:transfer_example:flag}, 
biasing the model to make the flag in the image an 
American flag,
all images for the base prompt for both models
do not feature an American flag but images for the manipulated prompt
feature an American flag. For the original model,
\imageScoresDelta is 91 percentage points and it is 95 points
for the transfer model. 
This can be explained by the fact that base images for the transfer 
model feature a different (not American) flag, while base images 
for the original model feature a blank flag.
In this case, we actually see slightly \emph{more} manipulation 
success in the transfer attack than in the original.

\begin{table*}[h]
    \centering
    \caption{Prompt and image quality for experiments with Stable Diffusion.
    Questions in each column are described in 
    \cref{sec:appendix:quality}.
    \emph{subset} are VL results
    on the human evaluation dataset.
    Base vs. manipulated prompt quality show small differences ($\Delta$ rows), 
    indicating minimal manipulation side effects.
    \pdouble$\pval<0.01$ and $*$ $\pval<0.05$, as measured by
    a paired t-test or Wilcoxon signed-rank test for non-normal samples.
    }
    \label{tab:appendix:quality:sd}
    \small
    \begin{tabular}{@{}r@{\hspace{0.5em}}r@{\hspace{0.5em}}r@{\hspace{0.25em}}
        r@{\hspace{0.25em}}r@{\hspace{0.25em}}r@{\hspace{0.5em}}
        r@{\hspace{0.25em}}r@{\hspace{0.25em}}r@{\hspace{0.25em}}
        r@{\hspace{0.75em}}r@{\hspace{0.25em}}
        r@{\hspace{0.25em}}r@{\hspace{0.5em}}r@{\hspace{0.25em}}r@{}}
    \toprule
    \midrule
        & & \multicolumn{8}{c}{Image Quality} & \multicolumn{5}{c}{Prompt Quality} \\
    \cmidrule(lr){3-10} \cmidrule(lr){11-15}
        & & \multicolumn{4}{c}{positive $\uparrow$} & \multicolumn{4}{c}{negative $\downarrow$} 
        & \multicolumn{3}{c}{positive $\uparrow$} & \multicolumn{2}{c}{negative $\downarrow$} \\
        \cmidrule(lr){3-6} \cmidrule(lr){7-10} \cmidrule(lr){11-13} \cmidrule(lr){14-15}
        judge & type 
        & \rotatebox{90}{on topic} & \rotatebox{90}{satisfied} 
        & \rotatebox{90}{clear} & \rotatebox{90}{like} 
        & \rotatebox{90}{artifact} & \rotatebox{90}{error} 
        & \rotatebox{90}{synthetic} & \rotatebox{90}{wrong} 
        & \rotatebox{90}{likely} & \rotatebox{90}{sense} 
        & \rotatebox{90}{pos} 
        & \rotatebox{90}{synthetic} & \rotatebox{90}{neg}\\
    \midrule
        \multirow{3}{*}{VL} & base& 71.90~ & 79.37~ & 74.85~ & 80.48~ & 9.40~ & 24.02~ & 5.20~ & 17.05~ & 41.05~ & 89.68~ & 72.69~ & 10.65~ & 9.48~ \\ 
         & man. & 80.27~ & 77.12~ & 70.43~ & 77.34~ & 10.57~ & 24.87~ & 4.83~ & 19.80~ & 45.73~ & 81.96~ & 65.65~ & 10.48~ & 5.72~ \\ 
         & $\Delta$&8.38\pdouble &-2.25* &-4.41\pdouble &-3.14\pdouble &1.17~ &0.85~ &-0.37~ &2.75~ &4.67\pdouble &-7.72\pdouble &-7.04\pdouble &-0.17~ &-3.76\pdouble \\ 
 
    \midrule 
 
        \multirow{3}{*}{subset} & base& 57.93~ & 77.84~ & 76.88~ & 80.24~ & 7.19~ & 22.43~ & 4.23~ & 13.41~ & 41.43~ & 90.71~ & 73.62~ & 9.21~ & 9.03~ \\
        & man.& 81.07~ & 79.90~ & 76.77~ & 80.87~ & 10.18~ & 26.25~ & 4.31~ & 17.80~ & 47.32~ & 84.43~ & 61.98~ & 9.01~ & 4.63~ \\ 
         & $\Delta$& 23.14~ & 2.05~ & -0.12~ & 0.63~ & 2.99* & 3.82~ & 0.07~ & 4.38~ & 5.89~ & -6.28* & -11.64* & -0.20~ & -4.41~ \\ 
 
    \midrule 
 
        \multirow{3}{*}{human} & base& 67.52~ & 58.04~ & -- & -- & -- & -- & -- & -- & 45.54~ & 86.61~ & 88.39~ & 25.89~ & 31.25~ \\ 
         & man. & 79.35~ & 53.12~ & -- & -- & -- & -- & -- & -- & 31.25~ & 71.43~ & 85.71~ & 53.57~ & 35.42~ \\ 
         & $\Delta$& 11.83~ & -4.91~ & -- & -- & -- & -- & -- & -- & -14.29\pdouble & -15.18* & -2.68~ & 27.68\pdouble & 4.17~ \\
    \midrule
    \bottomrule
    \end{tabular}
\end{table*}

\begin{table*}[h]
    \centering
    \caption{Prompt and image quality for experiments with Flux,
    following the same format as those for Stable Diffusion
    in \cref{tab:appendix:quality:sd}.
    }
    \label{tab:appendix:quality:flux}
    \small
    \begin{tabular}{@{}r@{\hspace{0.5em}}r@{\hspace{0.5em}}r@{\hspace{0.25em}}
        r@{\hspace{0.25em}}r@{\hspace{0.25em}}r@{\hspace{0.5em}}
        r@{\hspace{0.25em}}r@{\hspace{0.25em}}r@{\hspace{0.25em}}
        r@{\hspace{0.75em}}r@{\hspace{0.25em}}
        r@{\hspace{0.25em}}r@{\hspace{0.5em}}r@{\hspace{0.25em}}r@{}}
    \toprule
        \midrule
            & & \multicolumn{8}{c}{Image Quality} & \multicolumn{5}{c}{Prompt Quality} \\
        \cmidrule(lr){3-10} \cmidrule(lr){11-15}
            & & \multicolumn{4}{c}{positive $\uparrow$} & \multicolumn{4}{c}{negative $\downarrow$} 
            & \multicolumn{3}{c}{positive $\uparrow$} & \multicolumn{2}{c}{negative $\downarrow$} \\
            \cmidrule(lr){3-6} \cmidrule(lr){7-10} \cmidrule(lr){11-13} \cmidrule(lr){14-15}
            judge & type 
            & \rotatebox{90}{on topic} & \rotatebox{90}{satisfied} 
            & \rotatebox{90}{clear} & \rotatebox{90}{like} 
            & \rotatebox{90}{artifact} & \rotatebox{90}{error} 
            & \rotatebox{90}{synthetic} & \rotatebox{90}{wrong} 
            & \rotatebox{90}{likely} & \rotatebox{90}{sense} 
            & \rotatebox{90}{pos} 
            & \rotatebox{90}{synthetic} & \rotatebox{90}{neg}\\
    \midrule
        \multirow{3}{*}{VL} & base& 70.37~ & 88.30~ & 84.52~ & 88.75~ & 7.86~ & 21.73~ & 4.30~ & 12.20~ & 41.11~ & 89.89~ & 73.04~ & 10.66~ & 9.42~ \\ 
         & man. & 80.15~ & 86.55~ & 79.39~ & 86.35~ & 10.02~ & 28.48~ & 4.76~ & 16.46~ & 46.70~ & 81.04~ & 64.38~ & 12.55~ & 6.21~ \\ 
         & $\Delta$&9.77\pdouble &-1.75~ &-5.14\pdouble &-2.40* &2.16* &6.75\pdouble &0.47~ &4.26\pdouble &5.59\pdouble &-8.86\pdouble &-8.67\pdouble &1.89\pdouble &-3.21\pdouble \\ 
 
 \midrule 
 
        \multirow{3}{*}{subset} & base& 61.84~ & 84.53~ & 84.51~ & 84.28~ & 7.69~ & 18.93~ & 3.68~ & 12.44~ & 40.53~ & 91.82~ & 69.11~ & 9.61~ & 10.56~ \\ 
         & man. & 80.37~ & 89.10~ & 82.94~ & 89.53~ & 5.74~ & 21.50~ & 2.87~ & 13.48~ & 44.55~ & 82.39~ & 60.87~ & 11.44~ & 7.51~ \\ 
         & $\Delta$& 18.53~ & 4.57~ & -1.57~ & 5.25~ & -1.94~ & 2.57~ & -0.81~ & 1.03~ & 4.02~ & -9.43* & -8.24~ & 1.83~ & -3.05~ \\ 
 
 \midrule 
 
                \multirow{3}{*}{human} & base& 66.85~ & 78.46~ & -- & -- & -- & -- & -- & -- & 50.89~ & 94.64~ & 92.86~ & 34.82~ & 35.42~ \\ 
         & man. & 84.26~ & 69.08~ & -- & -- & -- & -- & -- & -- & 44.64~ & 82.14~ & 96.43~ & 40.18~ & 36.11~ \\ 
         & $\Delta$& 17.41~ & -9.38* & -- & -- & -- & -- & -- & -- & -6.25* & -12.50\pdouble & 3.57~ & 5.36~ & 0.69~ \\
         
    \midrule
    \bottomrule
    \end{tabular}
\end{table*}
\begin{table}[h]
    \centering
    \caption{
    Manipulated vs. base
    prompt readability. We report Pearson correlation 
    between readability and manipulation success.
    Manipulated prompts generally read easier;
    manipulation success weakly correlates with readability.
    $\uparrow$ means a higher value indicates more readable
    and $\downarrow$ is the opposite. \pdouble$\pval<0.01$ and $*$ $\pval<0.05$.
    }
    \label{tab:appendix:quality:prompt:sim}
    \begin{tabular}{@{}r@{\hspace{1em}}r@{\hspace{1em}}r@{\hspace{1em}}r@{\hspace{1em}}r@{\hspace{1em}}r@{}}
    \toprule
    \midrule
         & FRES $\uparrow$ & FKGL $\downarrow$ & GFI $\downarrow$ & ARI $\downarrow$ & CLI $\downarrow$\\
    \midrule
        base & 53.19~ & 8.23~ & 10.50~ & 6.83~ & 9.47~ \\
        manip. & 63.72~ & 7.49~ & 9.98~ & 7.46~ & 8.69~ \\
        $\Delta$ & 10.53~ & -0.74~ & -0.53~ & 0.63~ & -0.77~ \\
        corr.   & 0.22\pdouble & -0.162$*$ & -0.22\pdouble & 0.00~ & -0.05~ \\
    \midrule
    \bottomrule
    \end{tabular}
\end{table}

\subsection{Quality}
\label{sec:appendix:quality}
In addition to \SysName prompts generating images with a target bias,
these images should remain comparable quality, so users may continue
taking the manipulator's prompt suggestions.
We evaluate this in \cref{sec:appendix:quality:image}.
\SysName prompts should also be comparable quality---both not giving away
their bias and generally high-quality---in order for victims to use
the suggested prompts in the first place.
We describe how we evaluate this in \cref{sec:appendix:quality:prompt}.

\subsubsection{Image quality} 
\label{sec:appendix:quality:image}
We evaluate image quality with various VL model queries,
that consist of the image and a question about this image,
similarly to how we evaluate the image score in accordance to 
the image policy. 
We use the same VL model as \SysName, Qwen 2.5 VL.
Positive attributes of images can be difficult to pinpoint---previous work
with human subjects asked questions about image quality 
and image appeal~\cite{Goring2023Quality, Li2024AGIQA}.
Other work tests additional metrics, such as number of people 
or presence of a storyline~\cite{Shaojing2014Quality}.
Because these metrics are much more dependent on image subject,
we instead focus on the general attributes.
We evaluate the positive attributes with the following:
``Is this image on topic for the category `\{\}'?'' (\textbf{on topic}),
``Are you satisfied with the quality of this image?'' (\textbf{satisfied}),
``Is it clear what is depicted in this image?'' (\textbf{clear}), and
``Do you like the quality of this image?'' (\textbf{like}).
We evaluate the change in quality with a paired t-test
or Wilcoxon signed-rank test (for non-normally distributed samples).

Previous work highlights 
the presence of artifacts in images~\cite{Li2024AGIQA}, 
noise~\cite{Wang2024Quality}, 
or incorrect elements~\cite{Li2024AGIQA, Wang2024Quality}.
Based on these negative attributes, we evaluate the following questions:
``Is there an error with this image?'' (\textbf{error}),
``How synthetic does this image appear?'' (\textbf{synthetic}), and
``Is there anything wrong with the image?'' (\textbf{wrong}).
These image metrics are not included in \SysName's process of producing
a manipulated prompt. However, if a manipulator wishes, 
she could include any of these or others.

Overall image-quality shifts are small 
(see \crefrange{tab:appendix:quality:sd}{tab:appendix:quality:flux}).
For both Stable Diffusion and Flux, we see a statistically significant
(albeit small) increase in how on topic the images are.
For Stable Diffusion, we see statistically significant but small 
($<5.5$ percentage points)
decreases in the other positive qualities 
(\textbf{satisfied}, \textbf{clear}, \textbf{like}).
On Flux, we see similarly small decreases for \textbf{clear} and \textbf{like}. 
Flux also shows small ($\leq 6.75$ percentage points) 
but statistically significant increases in negative qualities.
These metrics are 
not part of \SysName by default but can be 
added to the policy when quality constraints are important.

\paragraph{Human image judge} 
We also perform human evaluation for the first two questions on 
a subset of the experiments (\textbf{on topic} and \textbf{satisfied}).
Here, we again see an increase in how on topic the images are, both 
for Flux and Stable Diffusion. However, this change is not 
statistically significant.
On Flux, we see a 9.38 percentage point decrease in how satisfied
human evaluators are with the image.
Nonetheless, the overall satisfaction score is still high (69.08\%),
and this drop is modest compared to manipulation success of \SysName.

\takeaways{
    Image quality does not differ significantly between 
    \SysName manipulated prompts and base prompts, with changes on all 
    metrics below 10 percentage points.
    Human evaluation is generally in line with VL model evaluation of image 
    quality.
}

\subsubsection{Prompt quality} 
\label{sec:appendix:quality:prompt}
For prompt evaluation, we consider both positive and negative prompt 
attributes from the prompt policy. 
We evaluate them with queries to the same VL model, Qwen 2.5 VL, consisting of
the prompt and a question about a certain attribute.
Good attributes are judged with the following:
``Are you likely to use this prompt?'' (\textbf{likely}),
``Does this prompt make sense?''---part of \promptPolicyPos (\textbf{sense}),
and any questions used by \SysName that correspond
    to a positive attribute (i.e. remaining questions in \promptPolicyPos),
    generally used to determine that the question is on topic (\textbf{pos}).
Bad attributes are judged with
``How synthetic does this prompt appear?'' (\textbf{synthetic})
and any prompt questions used by \SysName that correspond to a negative 
attribute (i.e. \promptPolicyNeg), generally used to determine if 
there is bias present in the prompt (\textbf{neg}).
This is measured both by the VL model and by human annotators.
We again use either a paired t-test or Wilcoxon signed-rank test to evaluate
statistical significance.

We show the evaluation of prompts on various 
prompt attributes in 
\crefrange{tab:appendix:quality:sd}{tab:appendix:quality:flux}.
For both models, we see a statistically significant increase 
for \textbf{likely} and a statistically significant decrease in 
general negative attributes for the prompt.
However, we also see a statistically significant decrease
in the general positive attributes and how much the prompt makes sense
for both models and an increase in how synthetic a prompt appears for Flux.
All these changes, both positive and negative, remain below 9 percentage points.

\paragraph{Human prompt judge}
Human judgement was more negative than VL judgement.
The judges were less likely to use the prompt
($14.29$pp for Stable Diffusion and $6.25$ for Flux) and said the 
prompt made less sense
($15.18$pp for Stable Diffusion and $12.5$ for Flux).
Most significantly, there was a 27.68 percentage point increase in how 
synthetic the judges found prompts for Stable Diffusion.
The remaining changes were statistically insignificant. 
but harsher on other prompt attributes. 
Nonetheless, with a possible exception of how synthetic the prompt appears, 
the bias benefits to the manipulator 
likely outweigh the prompt possible quality degradation.
If concerned with perceived syntheticness; a manipulator might need to 
try multiple base prompts or introduce a screening step for manual review.

\takeaways{
    VL evaluation indicates base vs. \SysName prompt quality is similar.
    Human ratings mostly align but indicate filtering may be needed.
}

\subsubsection{Prompt readability} 
\label{sec:appendix:quality:readability}
We also evaluate prompts on existing text metrics. 
First, we use the Flesch Reading Ease Score \textbf{(FRES)}~\cite{kincaid1975FKGL}, 
which estimates readability with word, sentence, and syllable counts.
Higher scores indicate easier reading.
We also use metrics that estimate 
years of education necessary to read 
a text---Flesch-Kincaid Grade Level \textbf{(FKGL)}~\cite{kincaid1975FKGL},
Automated Readability Index \textbf{(ARI)}~\cite{kincaid1975FKGL},
Coleman-Liau Index \textbf{(CLI)}~\cite{coleman1975cli}, 
Gunning Fog Index \textbf{(GFI)}~\cite{gunning1968gfi}.
These metrics are supplementary to VL prompt quality.

When we evaluate prompts for readability,
we saw that on average, FRES went up,
indicating text was easier to read.
The estimated grade level went down
for FKGL, GFI, and CLI
but up for the ARI.
We also saw statistically significant but low correlation
between the change in FRE and 
negative correlation for FKGL and Gunning Fog (see~\cref{tab:appendix:quality:prompt:sim}).

\subsection{Image score ratio}
\label{sec:appendix:ratio}
\begin{figure}[h!]
    \centering
    \includegraphics[width=\linewidth]{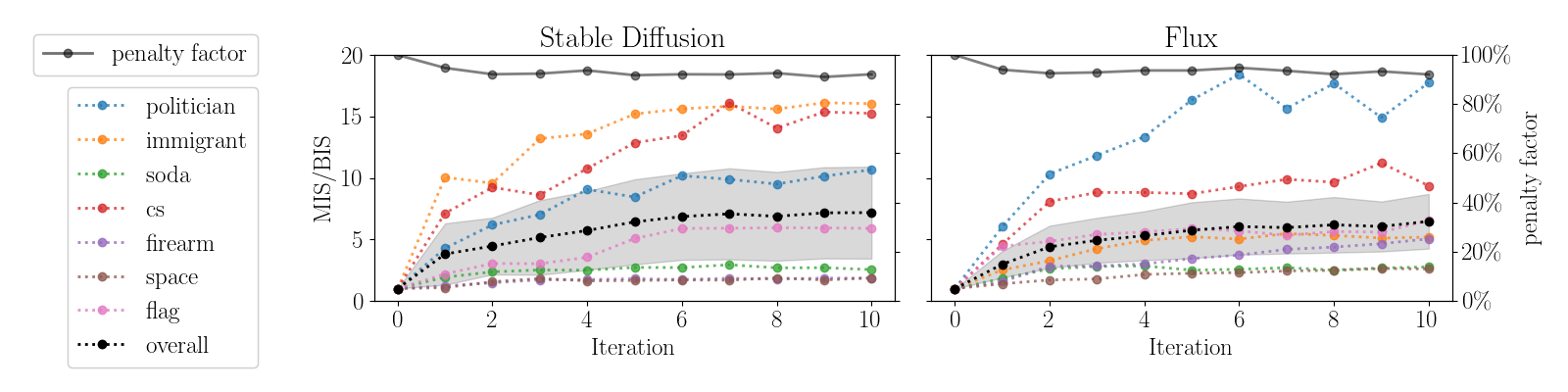}
    \caption{Average \imageScoresRatio for the best manipulated
    prompt by iteration, separated by category.
    Iteration 0, shows the base prompt.
    95\% confidence intervals are shaded for the overall average.
    }
    \label{fig:appendix:attack:success_over_time}
\end{figure}

In \cref{fig:evalres:attack:success_over_time},
we show the average best image score for a prompt tested 
at each iteration, or \imageScoresDelta.
In \cref{fig:appendix:attack:success_over_time}
we show an analogous figure but for \imageScoresRatio.
Manipulations for the category ``computer scientist''
perform better in terms of \imageScoresRatio than 
\imageScoresDelta, corroborated by \cref{tab:evalres:attack:success_auto}.
``Computer scientist'' is one of the worst performing categories
in terms of \imageScoresDelta. However, since \imageScoresBase
is so low for this category
(which means that the probability that the computer scientist
in an image is one of our targets, black or female,
is especially low for base prompts), the manipulation success in terms of 
ratio is greater.

\subsection{Implementation details}
\label{sec:appendix:details}

\begin{algorithm}[h!]
    \caption{Image or prompt judgements on individual 
    questions from the prompt and image policies.
    This returns the probability that the answer to the question is ``yes''.
    This is described in \cref{sec:method:individual}.}
    \label{alg:methods:score}
    \begin{algorithmic}
        \INPUT 
            $x$, the image (\generateImage[\manipulatedPrompt][\seed]) 
            or prompt (\basePrompt, \manipulatedPrompt)\\
            $q$, question from the image (\imagePolicy) 
            or prompt policies (\promptPolicyPos, \promptPolicyNeg)\\
            \postoks, the tokens for ``yes'', ``Yes'', and ``\xspace yes''\\
            \negtoks, the tokens for ``no'', ``No'', and ``\xspace no''\\
            \modelname, the VL model\\
        \OUTPUT
            \posprobs, the probability that the model would answer 
            question $q$ about $x$ affirmatively\\[8pt]

        \STATE $mask \gets \onehot[\postoks] \oplus \onehot[\negtoks]$
        \STATE $logits \gets \modelcall[x][q].logits[-1, -1, mask]$
        \STATE $\probs \gets \softmax[logits]$
        \STATE $\posprobs \gets \sum_{y \in \postoks} \probs[y]$
    \end{algorithmic}
\end{algorithm}

\begin{algorithm}[h!]
    \caption{
    Overall prompt scoring on the image and prompt policies.
    If the ratio between responses of the base and manipulated prompt 
    to a question in the prompt policy exceed a cutoff, it is 
    disqualified. Otherwise, deviations
    from the ratio are applied through \penaltyFunc 
    to reduce the penalty factor \penaltyFactor.
    The penalty is multiplied to the average score based on the image policy
    to get the overall score.
    This is described in \cref{sec:method:overall}.
    }
    \label{alg:methods:judge}
    \begin{algorithmic}
        \INPUT the base and manipulated prompts \basePrompt, \manipulatedPrompt \\
        image questions for the image policy \imagePolicy and weights \imagePolicyWeights \\
        prompt policy for positive attributes \promptPolicyPos and
        prompt policy for negative attributes \promptPolicyNeg \\
        constant maximum and minimum cutoffs \cutoffMin, \cutoffMax \\
        increasing (and concave) on valid inputs function \penaltyFunc\\
        image and prompt judging functions \scoreImageFuncName and \scorePromptFuncName, described in \cref{alg:methods:score}
        
        \OUTPUT image score \imScore, overall prompt score \score \\[8pt]

        \STATE $\disqualified 
        \gets \begin{array}[t]{@{}l@{}}
          \left(\exists \promptQuestion \in \promptPolicyPos ::
          \cutoffMin\leq\frac{\scorePromptFunc[\manipulatedPrompt][\promptQuestion]}{\scorePromptFunc[\basePrompt][\promptQuestion]}
            \right)~\vee
            \left(
            \exists \promptQuestion \in \promptPolicyNeg ::
            \cutoffMax\geq\frac{\scorePromptFunc[\manipulatedPrompt][\promptQuestion]}{\scorePromptFunc[\basePrompt][\promptQuestion]}
            \right)
            \end{array}$
        \STATE \COMMENT{check if prompt is disqualified}
        \IF{\disqualified}
            \STATE $\imScore, \score \gets 0, 0$ \COMMENT{prompt is disqualified}
        \ELSE
            \STATE $\imScore \gets 
                \frac{1}{\numSeeds}
                \sum\limits_{i=1}^{|\imagePolicy|} \imagePolicyWeights[i]
                \sum\limits_{\seed=1}^{\numSeeds} 
                \scoreImageFunc[\generateImage[\manipulatedPrompt][\seed]][\imagePolicy[i]]$

            \STATE $\penaltyFactor \gets 
            \begin{array}[t]{@{}l@{}}
              \left(\prod\limits_{\promptQuestion \in \promptPolicyPos}
                \penaltyFunc[\minFunc[\frac{\scorePromptFunc[\manipulatedPrompt][\promptQuestion]}{\scorePromptFunc[\basePrompt][\promptQuestion]}][1]]
                \right) * 
                \left(\prod\limits_{\promptQuestion \in \promptPolicyNeg}
                \penaltyFunc[\minFunc[\frac{\scorePromptFunc[\basePrompt][\promptQuestion]}{\scorePromptFunc[\manipulatedPrompt][\promptQuestion]}][1]]
                \right)
            \end{array}$
            \STATE \COMMENT{Apply penalty factor from prompt to image score}
            \STATE $\score \gets f*\imScore$
        \ENDIF
    \end{algorithmic}
\end{algorithm}

\begin{algorithm}[h!]
    \caption{
        The process by which the final manipulated prompt is found 
        through iteration and paraphrasing of the base prompt.
        This is described in \cref{sec:method:iterative}.
    }
    \label{alg:attack:iter}
    \begin{algorithmic}
        \INPUT the base prompt \basePrompt\\
        overall prompt scoring function \overallJudgeFuncName, described
        in \cref{alg:methods:judge}\\
        number of beams \numbeams\\
        \OUTPUT the final manipulated prompt \manipulatedPrompt\\[8pt]

        \STATE $\bestprompts \gets [\basePrompt]$
        \STATE $\allprompts \gets []$

        \FOR{$(\iter = 0; \iter < \maxiter; \iter++)$}
            \STATE $\iterprompts \gets []$
            \STATE \COMMENT{iterate over all beams}
            \FOR{$\bestprompt \in \bestprompts$} 
                \STATE \COMMENT{paraphrase best prompt of current beam}
                \FOR{$\prompt \in \paraphrase[\bestprompt]$}
                    \STATE \COMMENT{skip already seen prompts}
                    \IF{$\prompt \notin \allprompts \wedge \prompt \neq \basePrompt$}
                        \STATE \COMMENT{score current paraphrased prompt, as in \cref{alg:methods:judge}}
                        \STATE $\imScore, \score \gets \overallJudgeFunc[\basePrompt][\prompt]$
                        \STATE \COMMENT{check if prompt was disqualified}
                        \IF{$\score \neq 0$}
                            \STATE $\iterprompts \gets \iterprompts + [(\prompt, \score)]$
                            \STATE $\allprompts \gets \allprompts + [(\prompt, \score)]$
                        \ENDIF
                    \ENDIF
                \ENDFOR
            \ENDFOR
            \STATE $\iterprompts \gets \scoresort(\iterprompts)[-\numbeams:]$ 
            \COMMENT{get top \numbeams, based on score}
            \STATE $\iterScore \gets \frac{1}{k} \sum_{\prompt, \score \in \iterprompts} \score$
            \COMMENT{get current average score}
            \STATE $\bestScore \gets \frac{1}{|\bestprompts|} \sum_{\prompt, \score \in \bestprompts} \score$
            \COMMENT{get previous average score}
            \IF{$\iterScore < c \cdot \bestScore$}
                \STATE \COMMENT{if score dropped too much, find next iteration's best prompts from previously seen}
                \STATE $\bestprompts \gets \bestUnused[\allprompts]$
            \ELSE
                \STATE \COMMENT{otherwise, this iteration's best prompts are used}
                \STATE $\bestprompts \gets \iterprompts$
            \ENDIF
        \ENDFOR
        \STATE $\manipulatedPrompt \gets \scoresort(\allprompts)[-1,0]$
    \end{algorithmic}
\end{algorithm}

\subsubsection{Additional algorithms}
\label{sec:appendix:details:algs}
We provide additional details for the design of \SysName through
\crefrange{alg:methods:score}{alg:attack:iter}.
These algorithms provide detail the way \SysName makes individual
judgements (probability of a yes answer) on a prompt or image policy question 
(\cref{alg:methods:score}), how a prompt is judged in comparison
to the base prompt in accordance to the prompt and image policies
(\cref{alg:methods:judge}), and an overall view of the iterative process
(\cref{alg:attack:iter}).
The implementation for these algorithm is included in the 
supplementary material for this submission.

\subsubsection{Reproducibility}
\label{sec:appendix:details:reproducibility}
In this subsection, we describe details of our implementation
required for full reproducibility but not fully detailed in the main body 
of the paper. We also provide the code for our experiments in 
our supplementary material.

\paragraph{Image generation}
We use existing models in our paper, as described in their model cards.
We generate images of with dimensions $512\times512$ pixels,
with 28 inference steps, max sequence length of 512, and guidance 
scale 512 for Stable Diffusion and Flux. 
We pass in a generator, \verb|torch.manual_seed(seed)|,
for seeds from 0 to 7.
We generate 8 images for all prompts.
For our API model, GPT-Image, we use the provided API to get images of 
size $1024\times1024$ of quality ``medium''.

\paragraph{Paraphrasing}
When we get prompt paraphrases, we use use set the number of beams,
number of beam groups, and number of return sequences to the same value. We 
use a diversity penalty of one and a maximum length of 64 tokens.

\paragraph{Iteration}
In our iteration process, shown in high level in \cref{alg:attack:iter},
we use three beams. We begin by generating 10 paraphrases. 
We increase this by 10 if too many prompts are discarded during 
an iteration, so less valid prompts remain than the number of beams.
This can increase up to 30 paraphrases. 
If, at that point, too many prompts are discarded, 
we pull new best prompts from the best prompts that have not yet been
paraphrased, and set the number of paraphrases back to 10. 
We also pull new best prompts from the best prompts that have not yet been
paraphrased if the average score of the three best prompts at the end of the 
iteration is lower than the average score of the last three iterations
(if there have been less than three previous iterations,
we consider all previous iterations).
Our exact implementation is submitted in our supplementary material.

\paragraph{Judging prompts}
As described in \cref{sec:methods}, we compare prompts to the base prompt.
This can result in a prompt being disqualified, if the ratio of the 
scores is beyond a certain cutoff. For positive prompt attributes
(e.g. ``Is this prompt on topic for the category?'')
a prompt is disqualified if it scores $4\times$ lower than the base prompt.
For negative prompt attributes
(e.g. ``Does this prompt mention the gender of the subject?'')
a prompt is disqualified if it scores $2\times$ higher than the base prompt.
If the prompt is worse than the base prompt 
(lower for positive attributes, higher for negative attributes)
but is better than the cutoff, we adjust the penalty factor:
$f = f*\sqrt{1-(x-1)^2}$, where $x$ is the ratio of the 
prompt scores on the given question. This is shown in \cref{alg:methods:judge}.
The exact implementation is again provided in the supplementary material.

\subsubsection{Compute resources}
\label{sec:appendix:details:compute}

\paragraph{Hardware used}
All experiments were conducted on one or two NVIDIA RTX A5000 GPUs
and 128GB of CPU. We used our internal cluster.

\paragraph{Main contribution}
Each experiment---T2I model, base prompt, category, target---takes 12 GPU hours.
We ran 152 experiments per model, resulting in 3,648 GPU hours for the 
main body of our contributions.

Our main experiments with GPT Image are similar to the main 
contribution---the same judge and paraphrase models are loaded 
(which require GPUs),
but the T2I model is called via API, rather than loaded. 
We ran 14 experiments, which, in total, took 
under 168 GPU hours.

\paragraph{Alternative judge}
Our alternative judge experiments require no GPU, as they evaluate
existing images and prompts via an API model. Evaluating 152 experiments,
all experiments for a single T2I model, takes 2.5 CPU hours.
Both experiments with an alternative judge required 5 CPU hours.

\paragraph{Transfer attacks}
Transfer attacks are done by generating images for the base and manipulated
prompts via the API, then evaluating them. This evaluation is done
with a VL judge, which requires a GPU. Our experiments 
on a single NVIDIA RTX A5000 GPU took under 2 minutes.

\paragraph{Additional compute resources}
We performed preliminary experiments that are not reported in this paper.
These experiments also used NVIDIA RTX A5000 GPUs and used, in total,
less compute resources than the experiments that are reported in this paper.

\subsection{Time analysis}
\label{sec:appendix:details:time}
Here, we provide additional details to the time analysis in 
\cref{sec:discussion:analysis}.
Loading all models for an attack took 23.85 seconds for Stable Diffusion
and 20.09 seconds for Flux. 
Getting rephrasings took 0.54s, a prompt judgement took 0.04s, 
and an image judgement took 0.16s. 
Generating one image took 15.59s on Stable Diffusion and 21.59s
for Flux. 
The only steps of \SysName that must be done in sequence are
paraphrasing, image generation, and judgement, which means that the 
time per iteration is the sum of each step---giving us 16.33 second
iterations for Stable Diffusion and 22.33 second iterations for Flux.
If using an API model for image generation, 
loading required judge and paraphrasing models took 5.89s,
judgements and paraphrasing took 0.74s, and getting an image
took 14.93s, allowing for a 15.67s iteration.

\newpage

%% file: full.bib
@string{AIES = {AAAI/ACM Conference on AI, Ethics, and Society}}

@string{ACL = {Annual Meeting of the Association for Computational Linguistics}}

@string{CHI = {ACM Conference on Human Factors in Computing Systems}}

@string{CVPR = {IEEE Conference on Computer Vision and Pattern Recognition}}

@string{ECCV = {European Conference on Computer Vision}}

@string{FAACT = {ACM Conference on Fairness, Accountability, and Transparency}}

@string{ICCV = {IEEE/CVF International Conference on Computer Vision}}

@string{ICLR = {International Conference on Learning Representations}}

@string{ITVC = {IEEE Transactions on Visualization and Computer Graphics}}

@string{MM = {Proceedings of the 32nd ACM International Conference on Multimedia}}

@string{NEURIPS = {Conference on Neural Information Processing Systems}}

@string{NLPCC = {Natural Language Processing and Chinese Computing}}

@string{QoMEx = {International Workshop on Quality of Multimedia Experience}}

@string{TCSVT = {IEEE Transactions on Circuits and Systems for Video Technology}}

@string{TOIS = {ACM Transactions on Information Systems}}

@string{USESEC = {USENIX Security Symposium}}

@string{UIST = {Annual ACM Symposium on User Interface Software and Technology}}


%% file: refs.bib
@inproceedings{Arawjo_2024_chainforge, 
   title={ChainForge: A Visual Toolkit for Prompt Engineering and 
   `LLM Hypothesis Testing},
   url={http://dx.doi.org/10.1145/3613904.3642016},
   booktitle=CHI,
   author={I. Arawjo and C. Swoopes and P. Vaithilingam and 
   M. Wattenberg and E. L. Glassman},
   year={2024},
   month=may,
}

@INPROCEEDINGS{basu2023geo,
  author={A. Basu and R. V. Babu and D. Pruthi},
  booktitle=ICCV, 
  title={Inspecting the Geographical Representativeness of Images from Text-to-Image Models}, 
  year={2023},
  month=oct,
}

@inproceedings{bayramli2025diffusion,
    title = "Diffusion Models Through a Global Lens: Are They Culturally Inclusive?",
    author = {Z. Bayramli  and
      A. Suleymanzade and
      N. M. An and
      H. Ahmad and
      E. Kim and
      J. Park  and
      J. Thorne  and
      A. Oh},
    booktitle = {63\textsuperscript{rd} } # ACL,
    month = jul,
    year = 2025,
  pages     = {31137–31155},
}

@inproceedings{bender_etal_2021_stochastic,
  title     = {On the Dangers of Stochastic Parrots: Can Language Models Be Too Big?},
  author    = {Bender, Emily M. and Gebru, Timnit and McMillan-Major, Angelina and Shmitchell, Shmargaret},
  booktitle = FAACT,
  month     = mar,
  year      = {2021},
  pages     = {610-623},
}

@inproceedings{bianchi2023demographicbias,
  author = {F. Bianchi and P. Kalluri and E. Durmus and F. Ladhak and 
    M. Cheng and D. Nozza and T. Hashimoto and D. Jurafsky and 
    J. Zou and A. Caliskan},
  title = {Easily Accessible Text-to-Image Generation Amplifies Demographic Stereotypes at Large Scale},
  year = {2023},
  month = jun,
  booktitle = FAACT,
}

@misc{blackforestlabs2026flux,
  author = {Black Forest Labs},
  title = {FLUX.2 [klein]},
  howpublished={\url{https://huggingface.co/black-forest-labs/FLUX.2-klein-4B}},
  urldate = {2026-05-05}
}

@inproceedings{brade2023promptify,
  author = {S. Brade and B. Wang and M. Sousa and S. Oore and T. Grossman},
  title = {Promptify: Text-to-Image Generation through 
    Interactive Prompt Exploration with Large Language Models},
  year = {2023},
  month = may,
  booktitle={36\textsuperscript{th} } # UIST,
  pages = {1493 - 1504}
}

@inproceedings{chen2024tailoredvisions,
  title     = {Tailored Visions: Enhancing Text-to-Image Generation with Personalized Prompt Rewriting},
  author    = {Z. Chen and L. Zhang and F. Weng and L. Pan and Z. Lan},
  booktitle = CVPR,
  year      = {2024},
  month     = jun,
  pages = {7727-7736}
}

@inproceedings{chen2025t2i,
  title={T2i-copilot: A training-free multi-agent text-to-image system for enhanced prompt interpretation and interactive generation},
  author={Chen, Chieh-Yun and Shi, Min and Zhang, Gong and Shi, Humphrey},
  booktitle=ICCV,
  year={2025},
  pages = {19396-19405},
}

@inproceedings{chinchure2024tibet,
  title     = {TIBET: Identifying and Evaluating Biases in Text-to-Image Generative Models},
  author    = {A. Chinchure and P. Shukla and G. Bhatt and K. Salij and K. Hosanagar and L. Sigal and M. Turk},
  booktitle = {18\textsuperscript{th} } # ECCV,
  year      = {2024},
  month     = sep,
  pages = {429 - 446},
}

@article{coleman1975cli,
  title     = "A computer readability formula designed for machine scoring",
  author    = {M. Coleman and T. L. Liau},
  year      = 1968,
  journal = "Journal of Applied Psychology",
  volume = {60},
  pages = {283-284},
}

@inproceedings{cui2024phaseevo,
  title={PhaseEvo: Towards Unified Long-Context Prompt Optimization for Large Language Models},
  author={Wendi Cui and Jiaxin Zhang and Zhuohang Li and Hao Sun and Damien Lopez and Kamalika Das and Bradley A. Malin and Sricharan Kumar},
  booktitle={First Workshop on Long-Context Foundation Models @ ICML 2024},
  year={2024},
  month=jun,
}

@article{feng2023promptmagician,
  title     = {PromptMagician: Interactive Prompt Engineering for Text-to-Image Creation},
  author    = {Y. Feng and X. Wang and K. K. Wong and S. Wang and Y. Lu and M. Zhu and B. Wang and W. Chen},
  journal   = ITVC,
  year      = {2024},
  month     = jan,
  pages = {295 - 305},
}

@inproceedings{fernando2024PromptBreeder,
  author = {C. Fernando and D. Banarse and H. Michalewski 
    and S. Osindero and T. Rockt\"{a}schel},
  title = {Promptbreeder: self-referential self-improvement via prompt evolution},
  year = {2024},
  booktitle ={41\textsuperscript{st} } # ICML,
  month=jul,
}

@article{friedman_nissenbaum_1996_bias,
  title   = {Bias in Computer Systems},
  author  = {B. Friedman and H. Nissenbaum},
  journal = TOIS,
  volume  = {14},
  number  = {3},
  month   = jul,
  year    = {1996},
  pages   = {330--347},
}

@article{gallegos-etal-2024-bias,
  title = "Bias and Fairness in Large Language Models: A Survey",
  author = "Gallegos, Isabel O.  and
    Rossi, Ryan A.  and
    Barrow, Joe  and
    Tanjim, Md Mehrab  and
    Kim, Sungchul  and
    Dernoncourt, Franck  and
    Yu, Tong  and
    Zhang, Ruiyi  and
    Ahmed, Nesreen K.",
  journal = "Computational Linguistics",
  volume = "50",
  number = "3",
  month = sep,
  year = "2024",
  publisher = "MIT Press",
  pages = "1097--1179"
}

@inproceedings{gao_2024_align,
  author = {G. Gao and T. Taymanov and E. Salinas and P. Mineiro and D. Misra},
  title = {Aligning LLM agents by learning latent preference from user edits},
  year = {2024},
  booktitle = {38\textsuperscript{th} } # NEURIPS,
  month= may,
}

@InProceedings{Garcia_2023_CVPR,
  author = {Garcia, Noa and Hirota, Yusuke and Wu, Yankun and Nakashima, Yuta},
  title = {Uncurated Image-Text Datasets: Shedding Light on Demographic Bias},
  booktitle = CVPR,
  month = {June},
  year = {2023},
  pages = {6957-6966}
}

@inproceedings{goloujeh2024promptjourney,
  author = {A. Mahdavi Goloujeh and A. Sullivan and B. Magerko},
  title = {Is It {AI} or Is It Me? 
    Understanding Users’ Prompt Journey with Text-to-Image Generative AI Tools},
  year = {2024},
  month = may,
  series = CHI,
  pages = {1-13},
}

@inproceedings{Goring2023Quality,
  author={S. Göring and R. R. Ramachandra Rao 
    and R. Merten and A. Raake},
  booktitle={15\textsuperscript{th} } # QoMEX, 
  title = {Appeal and quality assessment for AI-generated images},
  year={2023},
  month=jun,
  pages={115-118}
}

@book{gunning1968gfi,
  title     = "The technique of clear writing",
  author    = {R. Gunning},
  year      = 1968,
  publisher = "McGraw-Hill",
}

@ARTICLE{guo2025PrompTHis,
  author={Y. Guo and H. Shao and C. Liu and K. Xu and X. Yuan},
  journal={IEEE Transactions on Visualization and Computer Graphics}, 
  title={PrompTHis: Visualizing the Process and 
    Influence of Prompt Editing During Text-to-Image Creation}, 
  year={2025},
  volume={31},
  number={9},
  pages={4547-4559},
  month=jun
}

@article{hao2023optimizing,
  title={Optimizing prompts for text-to-image generation},
  author={Hao, Yaru and Chi, Zewen and Dong, Li and Wei, Furu},
  journal=NEURIPS,
  volume={36},
  pages={66923--66939},
  year={2023}
}

@inproceedings{huang2025implicitbiasinjection,
  title     = {Implicit Bias Injection Attacks against Text-to-Image Diffusion Models},
  author    = {Huang, Huayang and Jin, Xiangye and Miao, Jiaxu and Wu, Yu},
  booktitle = CVPR,
  year      = {2025},
  pages     = {28779--28789},
}

@InProceedings{Kim_2025_CVPR,
  author = {Kim, Eunji and Kim, Siwon and Park, Minjun and Entezari, Rahim and Yoon, Sungroh},
  title = {Rethinking Training for De-biasing Text-to-Image Generation: Unlocking the Potential of Stable Diffusion},
  booktitle = CVPR,
  month = {June},
  year = {2025},
  pages = {13361-13370}
}

@article{kincaid1975FKGL,
  title     = "Derivation of new readability formulas 
    (automated readability index, fog count, and flesch reading ease formula) 
    for Navy enlisted personnel",
  author    = {J. P. Kincaid and R. P. Fishburne and 
    R. L. Rogers and B. S. Chissom},
  year      = 1975,
  month = feb,
  booktitle = "Naval Air Station Memphis: Chief of Naval Technical Training. Research Branch Report",
  pages = {8-75}
}

@article{Li2024AGIQA,
  author={C. Li and Z. Zhang and H. Wu and W. Sun and X. Min and X. Liu and 
    G. Zhai and W. Lin},
  journal=TCSVT, 
  title={AGIQA-3K: An Open Database for AI-Generated Image Quality Assessment}, 
  year={2024},
  month=sep,
  pages={6833 - 6846},
}

@inproceedings{Li2024PromptTist,
  author = {W. Li and J. Wang and X. Zhang},
  title = {PROMPTIST: Automated Prompt Optimization for
    Text-to-Image Synthesis},
  year = {2024},
  month = nov,
  booktitle = NLPCC,
}

@inproceedings{lin2025llm,
  title={{LLM} whisperer: {An} inconspicuous attack to bias {LLM} responses},
  author={W. Lin and A. Gerchanovsky and O. Akgul and L. Bauer and M. Fredrikson and Z. Wang},
  booktitle=CHI,
  pages={1--24},
  year={2025},
  
}

@inproceedings{luccioni_2023_neurips,
  title     = {Stable Bias: Evaluating Societal Representations in Diffusion Models},
  author    = {Luccioni, Sasha and Akiki, Christopher and Mitchell, Margaret and Jernite, Yacine},
  booktitle = NEURIPS,
  year      = {2023},
  pages={56338 - 56351},
}

@misc{midjourney2025describe,
  author       = {{Midjourney}},
  title        = {Describe},
  howpublished = {\url{https://docs.midjourney.com/hc/en-us/articles/32497889043981-Describe}},
  note         = {Accessed: 2026-01-17}
}

@article{mishra2025promptaid,
  title={PromptAid: Prompt Exploration, Perturbation, Testing and Iteration using Visual Analytics for Large Language Models}, 
  author={Aditi Mishra and Utkarsh Soni and Anjana Arunkumar and Jinbin Huang and Bum Chul Kwon and Chris Bryan},
  year={2025},
  month=oct,
  journal=ITVC,
  pages     = {969--974},
}

@inproceedings{naik2023socialbiases,
  title     = {Social Biases through the Text-to-Image Generation Lens},
  author    = {Naik, Ranjita and Nushi, Besmira},
  booktitle = AIES,
  year      = {2023},
  doi       = {10.1145/3600211.3604711},
  pages = {786 - 808},
}

@inproceedings{naseh2025backdooring,
  title = {Backdooring Bias {($B^2$)} into Stable Diffusion Models},
  author = {A Naseh and J. Roh and E. Bagdasarian and A. Houmansadr},
  booktitle = {34\textsuperscript{th} } # USESEC,
  year = 2025,
  month = aug,
  pages={977 - 996},
}

@article{newman2012truthiness,
  title     = {Nonprobative Photographs (or Words) Inflate Truthiness},
  author    = {Newman, Eryn J. and Garry, Maryanne and Bernstein, Daniel M. and Kantner, Justin and Lindsay, D. Stephen},
  journal   = {Psychonomic Bulletin \& Review},
  year      = {2012},
  volume    = {19},
  number    = {5},
  pages     = {969--974},
  doi       = {10.3758/s13423-012-0292-0}
}

@misc{openai2023dalle3,
  author       = {{OpenAI}},
  title        = {DALL{\textperiodcentered}E 3},
  howpublished = {\url{https://openai.com/index/dall-e-3/}},
  note         = {Accessed: 2026-01-17}
}

@misc{openai2025gptimage1,
  author       = {{OpenAI}},
  title        = {{GPT Image 1.5}},
  howpublished = {\url{https://developers.openai.com/api/docs/models/gpt-image-1.5}},
  year         = {2025},
  note         = {Accessed: 2026-05-05}
}

@misc{openai2025chat,
  author       = {{OpenAI}},
  title        = {{ChatGPT}},
  howpublished = {\url{https://chatgpt.com/}},
  year         = {2025},
  note         = {Accessed: 2026-05-05}
}

@misc{openai2025gpt41,
  author       = {{OpenAI}},
  title        = {{GPT 4.1}},
  howpublished = {\url{https://developers.openai.com/api/docs/models/gpt-4.1}},
  year         = {2025},
  note         = {Accessed: 2026-05-05}
}

@online{microsoft_copilot_imagine,
  author  = {{Microsoft}},
  title   = {Microsoft Copilot: Imagine},
  url     = {https://copilot.microsoft.com/imagine},
  urldate = {2026-01-27}
}

@InProceedings{Park_2025_ICCV,
  author = {Park, Jeonghoon and Lee, Juyoung and Chung, Chaeyeon and Lee, Jaeseong and Choo, Jaegul and Gu, Jindong},
  title = {Fair Generation without Unfair Distortions: Debiasing Text-to-Image Generation with Entanglement-Free Attention},
  booktitle = ICCV,
  month = {October},
  year = {2025},
  pages = {17567-17576}
}

@misc{qwen2025blog,
    title = {Qwen2.5-VL},
    url = {https://qwen.ai/blog?id=qwen2.5-vl},
    author = {Qwen Team},
    month = {January},
    year = {2025}
}

@misc{raza2025promptingaway,
  title   = {Prompting Away Stereotypes? {Evaluating} Bias in Text-to-Image Models for Occupations},
  author  = {Raza, Shaina and Powers, Maximus and Saha, Partha Pratim and Raza, Mahveen and Qureshi, Rizwan},
  year    = {2025},
  howpublished     = {\url{https://arxiv.org/abs/2509.00849}}
}

@inproceedings{seshadri-etal-2024-bias,
  title = "The Bias Amplification Paradox in Text-to-Image Generation",
  author = "Seshadri, Preethi  and
    Singh, Sameer  and
    Elazar, Yanai",
  booktitle = "Proceedings of the 2024 Conference of the North American Chapter of the Association for Computational Linguistics: Human Language Technologies (Volume 1: Long Papers)",
  month = jun,
  year = "2024",
  publisher = "Association for Computational Linguistics",
  pages = {6367–6384},
}

@INPROCEEDINGS{Shaojing2014Quality,
  author={Fan, Shaojing and Ng, Tian-Tsong and Herberg, Jonathan S. and Koenig, Bryan L. and Tan, Cheston Y.-C. and Wang, Rangding},
  booktitle=CVPR, 
  title={An Automated Estimator of Image Visual Realism Based on Human Cognition}, 
  year={2014},
  month=jun,
  pages={4201-4208},
}

@article{shin2020autoprompt,
  title={Autoprompt: Eliciting knowledge from language models with automatically generated prompts},
  author={Shin, Taylor and Razeghi, Yasaman and Logan IV, Robert L and Wallace, Eric and Singh, Sameer},
  journal={arXiv preprint arXiv:2010.15980},
  year={2020}
}

@misc{Siow:2025:Bart,
  author = {E. Siow},
  title = {{BART} paraphrase model (large)},
  howpublished={\url{https://huggingface.co/eugenesiow/bart-paraphrase}},
  urldate = {2025-12-2025}
}

@misc{stabilityai2026stablediffusion,
  author = {StabilityAI},
  title = {Stable Diffusion 3.5 Medium},
  howpublished={\url{https://huggingface.co/stabilityai/stable-diffusion-3.5-medium}},
  urldate = {2026-05-05}
}

@misc{su2023promptmindrefinepersonalized,
  title={Prompt Your Mind: 
    Refine personalized text prompts within your mind}, 
  author={Guinan Su and Yanwu Yang and Jie Guo},
  year={2023},
  month=nov,
  url={https://arxiv.org/abs/2311.05114}, 
}

@inproceedings{xiang2025promptsculptor,
  title={Promptsculptor: Multi-agent based text-to-image prompt optimization},
  author={Xiang, Dawei and Xu, Wenyan and Chu, Kexin and Ding, Tianqi and Shen, Zixu and Zeng, Yiming and Su, Jianchang and Zhang, Wei},
  booktitle={Proceedings of the 2025 Conference on Empirical Methods in Natural Language Processing: System Demonstrations},
  pages={774--786},
  year={2025}
}

@inproceedings{Wang2024PromptCharm,
  author = {Z. Wang and Y. Huang and D. Song and L. Ma and T. Zhang},
  title = {PromptCharm: Text-to-Image Generation through Multi-modal 
    Prompting and Refinement},
  year = {2024},
  month = may,
  booktitle = CHI,
}

@inproceedings{Wang2024Quality,
  author = {P. Wang and W. Sun and Z. Zhang and J. Jia and Y. Jiang and 
   Z. Zhang and X. Min and G. Zhai},
  title = {Large Multi-modality Model Assisted AI-Generated Image Quality Assessment},
  year = {2024},
  month = oct,
  booktitle = {32\textsuperscript{nd} } # MM,
  pages = {7803 - 7812}
}

@inproceedings{Wang2026Promptimizer,
  author = {L. Wang and K. Yurechko and A. X. Zhang},
  title = {Promptimizer: User-Led Prompt Optimization 
    for Personal Content Classification},
  year = {2026},
  month = may,
  booktitle = CHI,
}

@misc{wind2023prompts,
  title={{Top 40 useful prompts for Stable Diffusion XL}},
  author={D. Wind},
  year=2023,
  month=dec,
  howpublished={\url{https://medium.com/phygital/top-40-useful-prompts-for-stable-diffusion-xl-008c03dd0557}}
}

@inproceedings{yang2024large,
  title={Large Language Models as Optimizers},
  author={C. Yang and X. Wang and Y. Lu and 
    H. Liu and Q. V. Le and D. Zhou and X. Chen},
  booktitle=ICLR,
  year={2024},
  month=jan,
}

@inproceedings{zamfirescu2023johnny,
  author    = {J. D. Zamfirescu{-}Pereira and Richmond Y. Wong and Bjoern Hartmann and Qian Yang},
  title     = {Why {Johnny} Can't Prompt: {How} Non-{AI} Experts Try (and Fail) to Design {LLM} Prompts},
  booktitle = CHI,
  pages     = {437:1--437:21},
  publisher = {ACM},
  year      = {2023},
}

@InProceedings{Zhang_2023_ICCV,
  author = {Zhang, Cheng and Chen, Xuanbai and Chai, Siqi and Wu, Chen Henry and Lagun, Dmitry and Beeler, Thabo and De la Torre, Fernando},
  title = {ITI-GEN: Inclusive Text-to-Image Generation},
  booktitle = ICCV,
  month = {October},
  year = {2023},
  pages = {3969-3980}
}

@article{standing1973learning10000,
  title        = {Learning 10,000 pictures},
  author       = {Standing, Lionel},
  journal      = {Quarterly Journal of Experimental Psychology},
  year         = {1973},
  url          = {https://doi.org/10.1080/14640747308400340}
}

@article{newman2024,
  title        = {Misinformed by images: How images influence perceptions of truth and what can be done about it},
  author       = {Newman, Eryn J. and Schwarz, Norbert},
  journal      = {Current Opinion in Psychology},
  year         = {2024},
  url          = {https://doi.org/10.1016/j.copsyc.2023.101778}
}
